\documentclass[twoside,11pt]{article}

\usepackage{jmlr2e}

\usepackage{graphicx}
\usepackage{float}
\usepackage{multirow}
\usepackage{booktabs}
\usepackage{dcolumn}
\usepackage{makecell}
\usepackage{subcaption}
\usepackage{amssymb}
\usepackage{pifont}
\usepackage{amsmath}
\usepackage[table]{xcolor}
\usepackage{setspace}
\usepackage[misc]{ifsym}

\usepackage{hyperref}

\newcommand{\greencheck}{{\color{green}\ding{52}}}
\newcommand{\orangecheck}{{\color{orange}\ding{52}}}
\newcommand{\redcross}{{\color{red}\ding{56}}}

\ShortHeadings{AutoML Meets Time Series Regression}{Z. Xu, W. Tu and I. Guyon}
\firstpageno{1}

\begin{document}

\title{AutoML Meets Time Series Regression \\
{\large Design and Analysis of the  AutoSeries Challenge}}

\author{\name Zhen Xu \email xuzhen@4paradigm.com \\
       \addr 4Paradigm, China\\
       \AND
       \name Wei-Wei Tu \email tuweiwei@4paradigm.com \\
       \addr 4Paradigm, China\\
       \AND
       \name Isabelle Guyon \email guyon@chalearn.org \\
       \addr LISN CNRS/INRIA, France\\
       University Paris-Saclay, France\\
       ChaLearn, USA}

% \editor{Kevin Murphy and Bernhard Sch{\"o}lkopf}

\maketitle

\begin{abstract}%   <- trailing '%' for backward compatibility of .sty file
Analyzing better time series with limited human effort is of interest to academia and industry. Driven by business scenarios, we organized the first Automated Time Series Regression challenge (AutoSeries) for the WSDM Cup 2020. We present its design, analysis, and post-hoc experiments. The code submission requirement precluded participants from any manual intervention, testing automated machine learning capabilities of solutions, across many datasets, under hardware and time limitations. We prepared 10 datasets from diverse application domains (sales, power consumption, air quality, traffic, and parking), featuring missing data, mixed continuous and categorical variables, and various sampling rates.  Each dataset was split into a training and a test sequence (which was streamed, allowing models to continuously adapt). The setting of ``time series regression'', differs from classical forecasting in that covariates at the present time are known. 
Great strides were made by participants to tackle this AutoSeries problem, as demonstrated by the jump in performance from the sample submission, and post-hoc comparisons with AutoGluon. Simple yet effective methods were used, based on feature engineering, LightGBM, and random search hyper-parameter tuning, addressing all aspects of the challenge.  Our post-hoc analyses revealed that providing additional time did not yield significant improvements.  The winners' code was open-sourced\footnote{\url{https://github.com/NehzUx/AutoSeries}}. 
\end{abstract}

\section{Introduction}
\label{sec:intro}
Machine Learning (ML) has made remarkable progress in the past few years in time series-related tasks, including time series classification, time series clustering, time series regression, and time series forecasting \cite{wang_tsc,lim_tsf}.To foster research in time series analysis, several competitions have been organized, since the onset of machine learning. These include the Santa Fe competition\footnote{\url{https://archive.physionet.org/physiobank/database/santa-fe/}}, the Sven Crone competitions\footnote{\url{http://www.neural-forecasting-competition.com/}}, several Kaggle comptitions including M5 Forecasting\footnote{\url{https://www.kaggle.com/c/m5-forecasting-accuracy}}, Web Traffic Time Series Forecasting\footnote{\url{https://www.kaggle.com/c/web-traffic-time-series-forecasting}}, to name a few. While time series forecasting remains a very challenging problem for ML, successes have been reported on problems of time series regression and classification in practical applications \cite{tan_tsr,wang_tsc}.
 
Despite these advances, switching domain, or even analysing a new dataset from the same domain, still requires considerable human engineering effort. To address this problem, recent research has been directed to Automated Machine Learning (AutoML) frameworks \cite{automl_book,yao_automl}, whose charter is to reduce human intervention in the process of rolling out machine learning solutions to specific tasks. AutoML approaches include designing (or meta-learning) generic reusable pipelines and/or learning machine architectures, fulfilling specific task requirements, and designing optimization methods devoid of (hyper-)parameter choices. To stimulate research in this area, we launched with our collaborators a series of challenges exploring various application settings\footnote{\url{http://automl.chalearn.org}, \url{http://autodl.chalearn.org}}, whose latest editions include the Automated Graph Representation Learning (AutoGraph) challenge at the KDD Cup AutoML track\footnote{\url{https://www.automl.ai/competitions/3}}, Automated Weakly Supervised Learning (AutoWeakly) challenge at ACML 2019\footnote{\url{https://autodl.lri.fr/competitions/64}}, Automated Computer Vision (AutoCV\cite{liu_autocv}) challenges at IJCNN 2019 and ECML PKDD 2019, etc. 

This paper presents the design and results of the Automated Time Series Regression (AutoSeries) challenge, one of the competitions of the WSDM Cup 2020 (Web Search and Data Mining conference) that we co-organized, in collaboration with 4Paradigm and ChaLearn. 

This challenge addresses ``time series regression'' tasks \cite{fpp3_book}. In contrast with ``strict'' forecasting problems in which {\em forecast} variable(s) $\mathbf{y}_t$ should be predicted from {\bf past} values {\bf only} (often $\mathbf{y}$ values alone), {\bf time series regression} seeks to predict $\mathbf{y}_t$ using {\bf past} $\{t-t_{min}, \cdots , t-1 \}$ AND {\bf present} $t$ values of one (or several) ``covariate'' {\em feature} time series $\{\mathbf{x}_t\}$\footnote{In some application domains (not considered in this paper), even {\bf future} $ \{ t+1, \cdots, t+t_{max} \}$) values of the covariates may be considered. An example would be ``simultaneous translation'' with a small lag.}.  Typical scenarios in which $\mathbf{x}_t$ is known at the time of predicting $\mathbf{y}_t$ include cases in which $\mathbf{x}_t$ values are scheduled in advance or hypothesized for decision making purposes. Examples include: {\em scheduled events} like upcoming sales promotions, {\em recurring events} like holidays, or {\em forecasts} obtained by external accurate simulators, like weather forecasts. This challenge addresses in particular {\bf multivariate} time series regression problems, in which $\mathbf{x}_t$ is a feature vector or a matrix of  {\em covariate} information, and $\mathbf{y_t}$ is a vector. The domains considered include air quality, sales, parking, and city traffic forecasting. Data are feature-based and represented in a ``tabular'' manner. The challenge was run with {\bf code submission} and the participants were evaluated on the Codalab challenge platform, without any human intervention, on five datasets in the feedback phase and five different datasets in the final ``private'' phased (with full blind testing of a single submission).

While future AutoSeries competitions might address other difficulties, this particular competition focused on the following 10 questions:
\begin{itemize}
    \item[Q1:] {\bf Beyond autoregression: Time series regression.} Do participants exploit covariates/features $\{ \mathbf x_t \}$ to predict $y_t$, as opposed to only past $y$?
    \item[Q2:] {\bf Explainability.} Do participants make an effort to provide an explainable model, e.g. by identifying the most predictive features in $\{ \mathbf x_t \}$?
    \item[Q3:] {\bf Multivariate/multiple time series.} Do participants exploit the joint distribution/relationship of various time series in a dataset?
    \item[Q4:] {\bf Diversity of sampling rates.} Can methods developed handle different sampling rates (hourly, daily, etc.)?
    \item[Q5:] {\bf Heterogeneous series length.} Can methods developed handle truncated series either at the beginning or the end?
    \item[Q6:] {\bf Missing data.} Can methods developed handle (heavily) missing data?
    \item[Q7:] {\bf Data streaming.} Do models update themselves according to newly acquired streaming test data (to be explained in Sec \ref{protocol})?
    \item[Q8:] {\bf Joint model and HP selection.} Can models select automatically learning machines and hyper-parameters?
    \item[Q9:] {\bf Transfer/Meta learning.} Are solutions provided generic and applicable to new domains or at least new datasets of the same domain?
    \item[Q10:] {\bf Hardware constraints.} Are computational/memory limitations observed?
\end{itemize}

\iffalse
\begin{itemize}
    \item How to automatically discover various kinds of information from other variables besides time feature?
    \item How to automatically extract useful features from different datasets of time series data?
    \item How to automatically handle relations among different time series?
    \item How to automatically handle both long and short time series data?
    \item How to automatically design effective neural network structures?
    \item How to automatically and efficiently select appropriate machine learning model and hyper-parameters?
    \item How to make the solution more generic, i.e., how to make it applicable for unseen datasets?
    \item How to keep the computational and memory cost acceptable? 
\end{itemize}
\fi

\section{Challenge Setting}

\subsection{Phases}
\label{sec:phases}

The AutoSeries challenge had three phases: a {\bf Feedback Phase}, a  {\bf Check Phase} and a  {\bf Private Phase}. In the Feedback Phase, five ``feedback datasets'' were provided to evaluate participants' AutoML models. The participants could read error messages in log files made available to them (e.g. if their model failed due to missing values) and obtain performance and ranking feedback on a leaderboard. When the Feedback Phase finished, five new ``private datasets'' were used in the Check Phase and the Private Phase. The Check Phase was a brief transition phase in which the participants submitted their models to the platform to verify whether the model ran properly. No performance information or log files were returned to them. Using a Check Phase is a particular feature of this challenge, to avoid disqualifying participants on the sole ground that their models timed out, used an excessive amount of memory, or raised another exception possible to correct without specific feedback on performance. Finally in the Private Phase, the participants submitted blindly their debugged models, to be evaluated by the same five datasets as in Check Phase.

As previously indicated, in addition to the five feedback datasets and five private datasets, two public datasets were provided for offline practice.  

\begin{figure}[h]
% \centering
\centering
\includegraphics[width=0.85\textwidth]{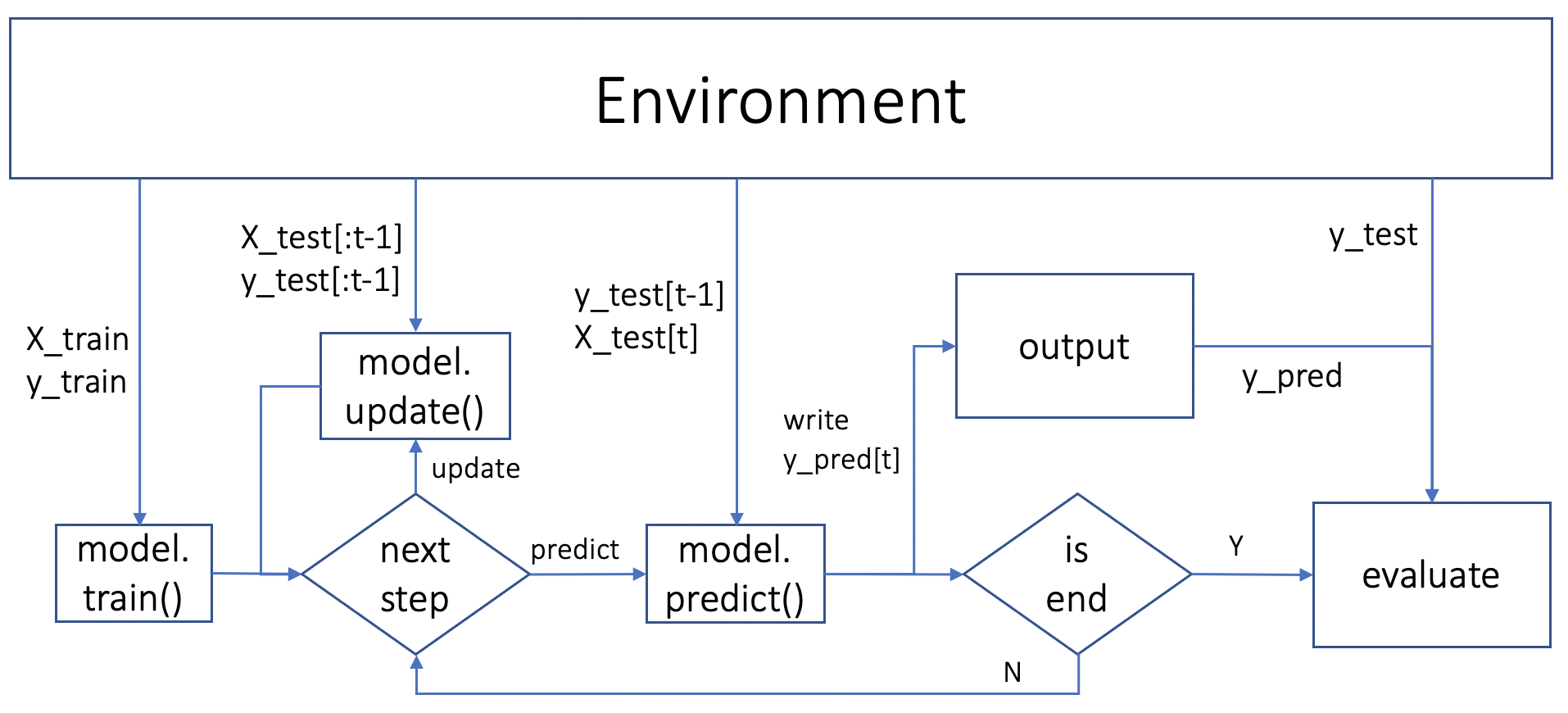}
\caption{ {\bf Challenge protocol.} \texttt{train}, \texttt{update}, and \texttt{predict} methods must be provided by participants. Such methods are under control of timers, omitted in the figure.}
\label{fig:protocol}
\end{figure}

\subsection{Protocol}
\label{protocol}

The AutoSeries challenge was designed based on real business scenarios, emphasizing {\bf automated machine learning} (AutoML) and {\bf data streaming}. First, as in other AutoML challenges, algorithms were evaluated on various datasets entirely hidden to the particpants, {\bf without any human intervention}. In other time series challenges, such as Kaggle's  Web Traffic Time Series Forecasting\footnote{\url{https://www.kaggle.com/c/web-traffic-time-series-forecasting}}), participants downloaded and explored past training data, and manually tuned features or models. The AutoSeries challenge forced the participants to design {\bf generic methods}, instead of developing {\it ad hoc} solutions. Secondly, test data were streamed such that at each time point $t$, historical information of past time steps $\mathbf{x}_{train}[:t-1]$, $\mathbf{y}_{train}[:t-1]$ and features of time $t$, X\_test[t] were available for predicting $\mathbf{y}_t$. In addition to the usual \texttt{train} and \texttt{predict} methods, the participants had to prepare a method \texttt{update}, together with a  strategy to update their model at an appropriate frequency, once fully trained on the training data. Updating too frequently might lead to run out of time; updating not frequently enough could result in missing recent useful information and performance degradation. The protocol is illustrated in Figure \ref{fig:protocol}. 

\subsection{Datasets}
The datasets from the Feedback Phase and final Private Phase are listed in Table \ref{tab:stats}. We purposely chose datasets from various domains, having a diversity of types of variables (continuous/categorical), number of series, noise level, amount of missing values, and sampling frequency (hourly, daily, monthly), and level of nonstationarity. Still, we eased the difficultly by including in each of the two phases datasets having some resemblance. 

\begin{figure}
    \centering
    \begin{subfigure}[b]{0.475\textwidth}
        \centering
        \includegraphics[width=\textwidth]{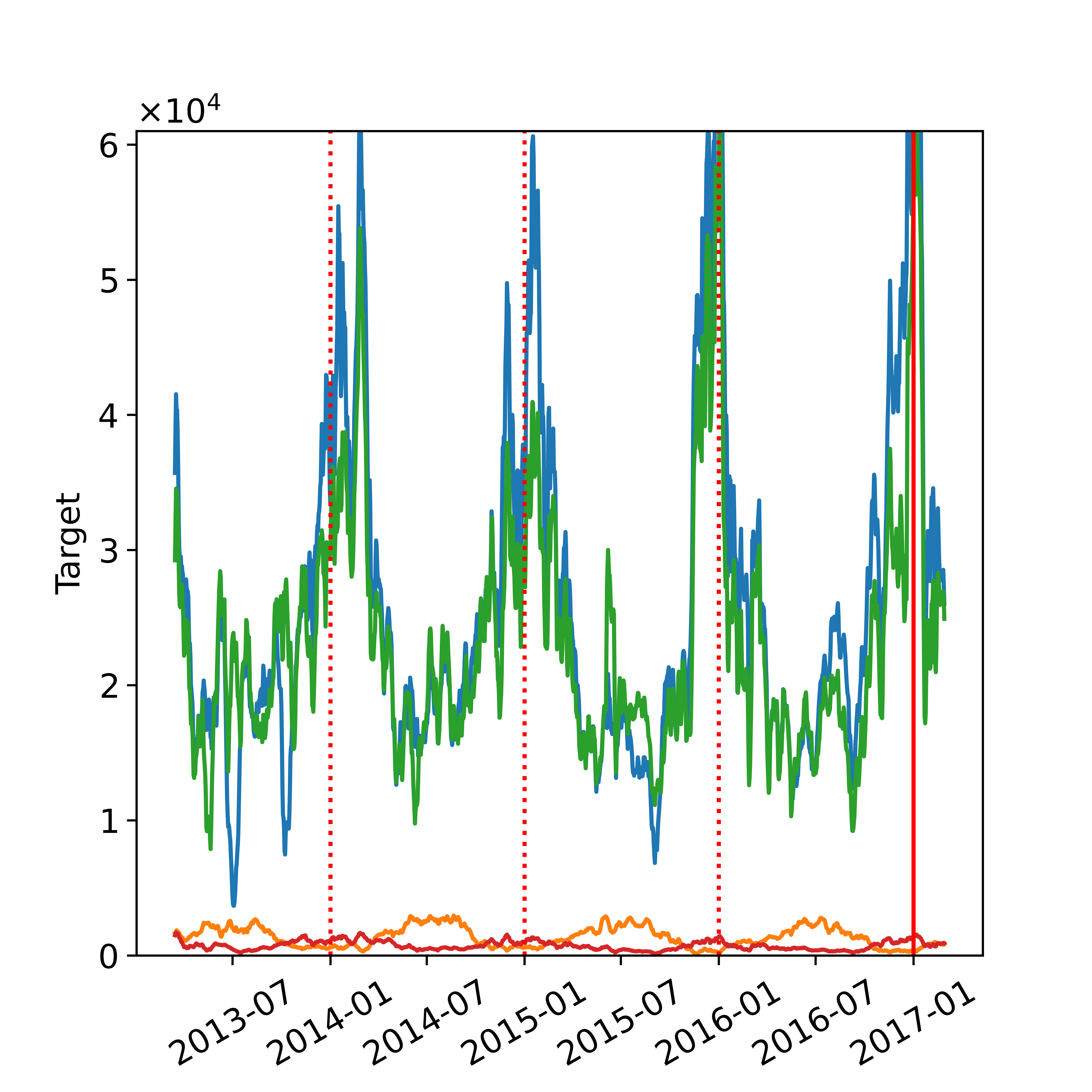}
        \caption[]{{\small Dataset fph2}} 
        \label{fig:fph2}
    \end{subfigure}
    % \hspace{0.1cm}
    \vspace{-0.15cm}
    \begin{subfigure}[b]{0.475\textwidth} 
        \centering 
        \includegraphics[width=\textwidth]{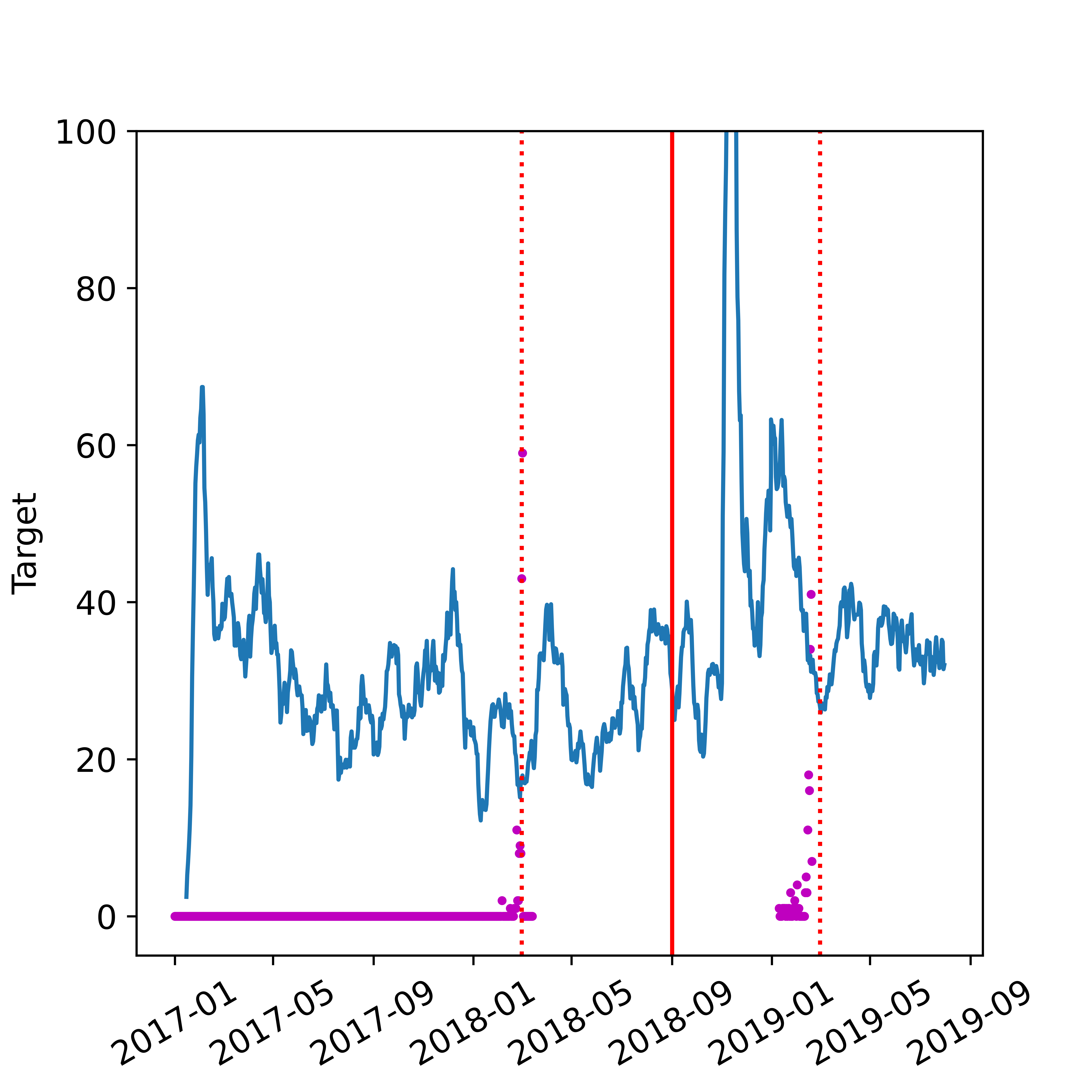}
        \caption[]{{\small Dataset pph3}} 
        \label{fig:pph3}
    \end{subfigure}
    \hspace{-0.1\textwidth}
    \begin{subfigure}[b]{0.485\textwidth}   
        \centering 
        \includegraphics[width=\textwidth]{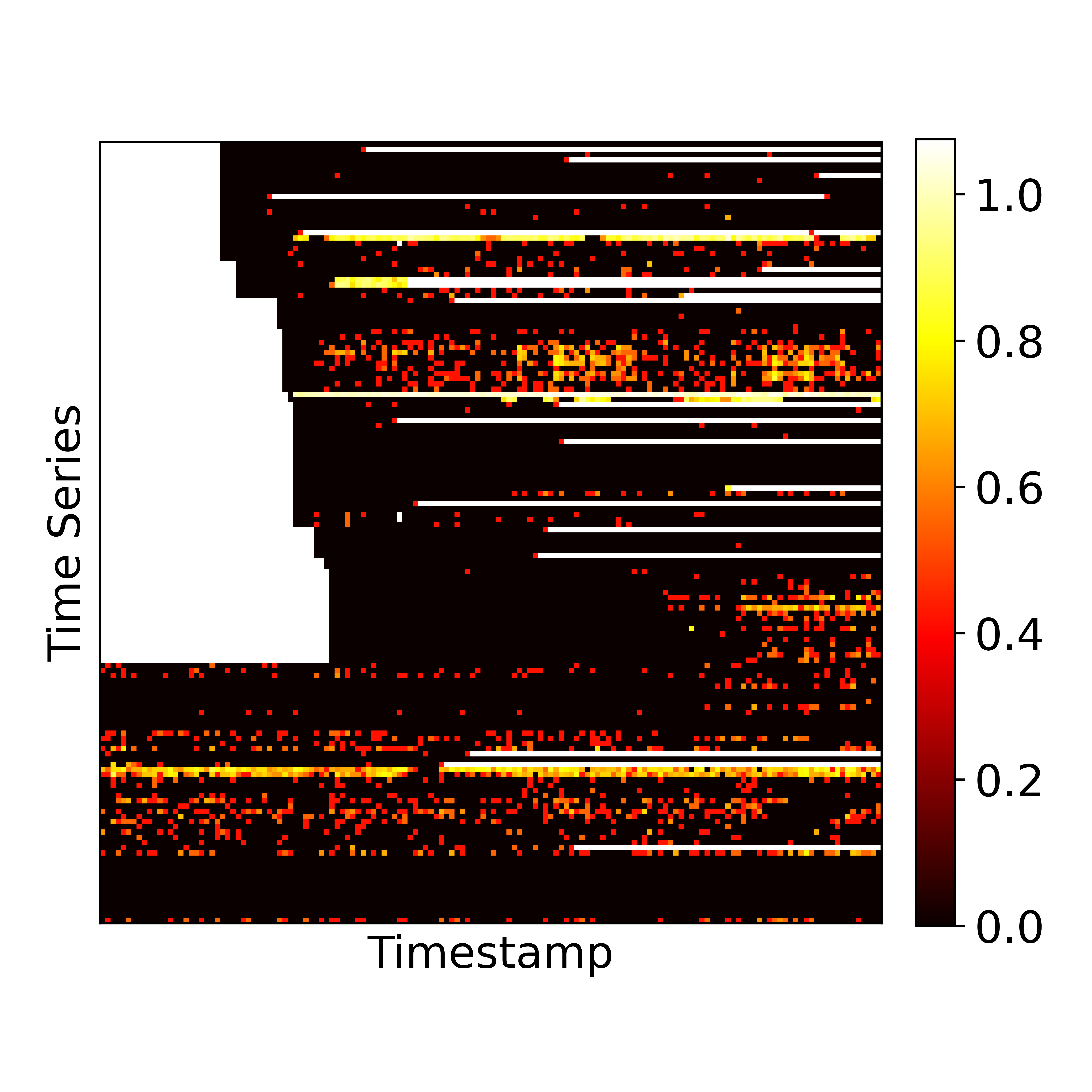}
        \caption[]{{\small Dataset fph5}}
        \label{fig:fph5}
    \end{subfigure}
    % \hspace{0.1\textwidth}
    \begin{subfigure}[b]{0.465\textwidth} 
        \centering 
        \includegraphics[width=\textwidth]{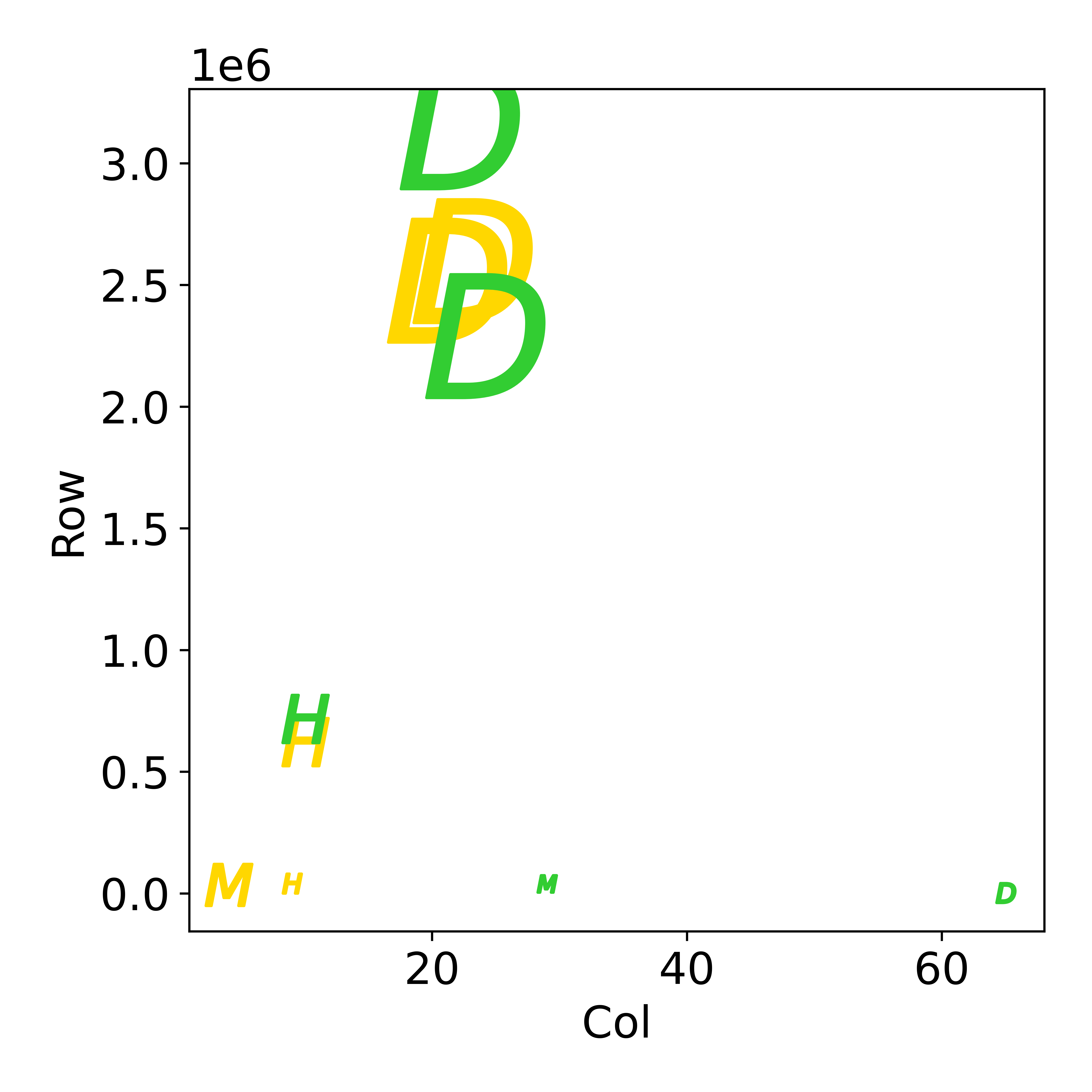}
        \caption[]{{\small Metadata}}    
        \label{fig:metadata}
    \end{subfigure}
    \caption{\footnotesize {\bf Dataset visualization.} \textbf{(a)} {\bf Sample visualization of dataset fph2.} Four time series (after smoothing and resampling). Training data is available until end of 2016 (red vertical solid line). Yearly preriodicity is indicated by dashed vertical lines to highlight seasonalities. One notices large differences in series amplitudes and patterns of seasonality. \textbf{(b)} {\bf Sample visualization of dataset pph3.} Two time series (after smoothing). Training data is available until end of 2018-08 (red vertical solid line). The purple time series suffers from missing values (certain items have zero sales most of the time). A clear trend exists in blue time series, unlike the example shown in (a).  \textbf{(c)} {\bf Heatmap visualization of dataset fph5.} White means missing value. Black means zero target value (sales). Several common issues can be observed: (1) many items don't sell most of the time; (2) presence of many missing values; (3) time series vary in lengths and are not aligned; (4) different time series have totally different scales. \textbf{(d)} {\bf All dataset metadata visualization.} X axis is the number of columns. Y axis is the number of rows. The symbol letter shape represents the time period: Monthly, Daily, or Hourly. The symbol color represents the phase: green for ``feedback'' and orange for ``private''. The symbol size represents the number of lines in the dataset.}
    \label{fig:datavis}
\end{figure}

Two types of tabular formats are commonly used: the ``wide format'' and ``long format''\footnote{\url{https://doc.dataiku.com/dss/latest/time-series/data-formatting.html}}. The wide format facilitates visualization and direct use with machine learning packages. It consists in one time record per line, with feature values (or series) in columns. However, for large number of features and/or missing values, the long format is preferred. In that format, a minimum of 3 columns are provided: (1) date and time (referred to as ``Main\_Timestamp''), (2) feature identifier (referred to as ``ID\_Key''), (3) feature value. Pivoting is an operation, which allows converting the wide format into the long format and vice-versa. From the long format, given one value of ID\_Key (or a set of ID\_Keys), a particular time series is obtained by ordering the feature values by Main\_Timestamp. In this challenge, since we address a time series regression problem, we add a fourth column (4) ``Label/Regression\_Value'' providing the target value, which must always be provided. A data sample in found in Table~\ref{tab:sample_fph2} and data visualizations in Figure \ref{fig:datavis}.

\begin{table}[ht!]
\caption{ {\bf Statistics of all 10 datasets.} \\
Sampling ``Period'' is indicated in (M)inutes, (H)ours, (D)ays. ``Row'' and ``Col'' are the total number of lines and columns, in the long format. Columns includes: Timestamp, (multiple) Id\_Keys, (multiple) Features, and Target. ``KeyNum'' is the number of Id\_Keys (called Id\_Key combination, e.g. in a sales problem Product\_Id and Store\_Id.) ``FeatNum'' indicates the number of features for each Id\_Key combination (e.g. for a given Id\_Key corresponding to a product in a given store, features include price, and promotion.) ``ContNum'' is the number of continuous features and ``CatNum'' is the number of categorical features; CatNum + ContNum = FeatNum.  ``IdNum'' means the number of unique Id\_Key combinations. One can verify that Col = 1 (timestamp) + KeyNum + FeatNum + 1 (target). ``Budget'' is the time in seconds that we allow participants' models to run.
}
\label{tab:stats}
\centering
\resizebox{\textwidth}{!}{
\begin{tabular}{|c|c|c|c|c|c|c|c|c|c|c|} 
\toprule
Dataset & Domain     & Period & Row     & Col & KeyNum & FeatNum & ContNum & CatNum & IdNum & Budget \\
\midrule
\rowcolor{green!40} fph1    & Power      & M      & 39470   & 29  & 1      & 26      & 26      & 0      & 2 & 1300   \\
\rowcolor{green!40} fph2    & AirQuality & H      & 716857  & 10  & 2      & 6       & 5       & 1      & 21 & 2000   \\
\rowcolor{green!40} fph3    & Stock      & D      & 1773    & 65  & 0      & 63      & 63      & 0      & 1 & 500    \\
\rowcolor{green!40} fph4    & Sales      & D      & 3147827 & 23  & 2      & 19      & 10      & 9      & 8904 &  3500   \\
\rowcolor{green!40} fph5    & Sales      & D      & 2290008 & 23  & 2      & 19      & 10      & 9      & 5209 & 2000   \\
\midrule
\rowcolor{yellow!60} pph1    & Traffic    & H      & 40575   & 9   & 0      & 7       & 4       & 3      & 1 & 1600   \\
\rowcolor{yellow!60} pph2    & AirQuality & H      & 721707  & 10  & 2      & 6       & 5       & 1      & 21 & 2000   \\
\rowcolor{yellow!60} pph3    & Sales      & D      & 2598365 & 23  & 2      & 19      & 10      & 9      & 6403 & 3500   \\
\rowcolor{yellow!60} pph4    & Sales      & D      & 2518172 & 23  & 2      & 19      & 10      & 9      & 6395 & 2000   \\
\rowcolor{yellow!60} pph5    & Parking    & M      & 35501   & 4   & 1      & 1       & 1       & 0      & 30 & 350    \\
 \bottomrule
\end{tabular}
}
\end{table}

\begin{table}[ht!]
\caption{ {\bf Sample data for dataset fph2.} A1 = timestamp. A2, A3, A4, A5, A7 = continuous features. A6 = categorical feature (hashed). A8, A9 = Id columns (hashed). Hashing is used for privacy. A10 = target.}
\label{tab:sample_fph2}
\centering

\resizebox{\textwidth}{!}{
\tiny
\begin{tabular}{|c|c|c|c|c|c|c|c|c|c|}
\toprule
A1 & A2 & A3 & A4 & A5 & A6 & A7 & A8 & A9 & A10 \\ 
\midrule
2013-03-01 00:00:00    & -2.3                   & 1020.8                 & -19.7                  & 0.0                    & -457...578   & 0.5                    & 657...216     & -731...089   & 13.0                    \\
2013-03-01 01:00:00    & -2.5                   & 1021.3                 & -19.0                  & 0.0                    & 511...667    & 0.7                    & 657...216     & -731...089   & 6.0                     \\
2013-03-01 02:00:00    & -3.0                   & 1021.3                 & -19.9                  & 0.0                    & 511...667    & 0.2                    & 657...216     & -731...089   & 22.0                    \\
$\cdots$& $\cdots$& $\cdots$& $\cdots$& $\cdots$& $\cdots$& $\cdots$& $\cdots$& $\cdots$& $\cdots$ \\
2017-02-28 19:00:00    & 10.3                   & 1014.2                 & -12.4                  & 0.0                    & 495...822    & 1.8                    & 784...375    & 156...398    & 27.0                    \\
2017-02-28 20:00:00    & 9.8                    & 1014.5                 & -9.9                   & 0.0                    & -286...752   & 1.5                    & 784...375    & 156...398    & 47.0                    \\
2017-02-28 21:00:00    & 9.1                    & 1014.6                 & -12.7                  & 0.0                    & -213...128   & 1.7                    & 784...375    & 156...398    & 18.0                   \\ 
\bottomrule

\end{tabular}
}
\end{table}

\subsection{Metrics}

The metric used to judge the participants is the \textsf{RMSE}. For each datasets, the participant's submissions are run in the same environment, and ranked according to the \textsf{RMSE} for each dataset. Then, an overall ranking is obtained from the average dataset rank, in a given phase. In post challenge analyses, we also used two other metrics: \textsf{SMAPE} and Correlation (\textsf{CORR}). The formulas are provided below. $y$ means ground truth target. $\hat y$ is the prediction. $\bar y$ is the mean. $N$ is total number of unique Id combinations (IdNum in Table \ref{tab:stats}) and $T$ is number of timestamps. For evaluation, these metrics are run on the test sequences only.

\begin{equation}
    \textsf{RMSE} = \sqrt{\frac{1}{NT} \sum_{n=1}^{N} \sum_{t=1}^{T} (y_{nt} - \hat y_{nt})^2}
\label{eq:rmse}
\end{equation}

\begin{equation}
    \textsf{SMAPE} = \frac{1}{NT}\sum_{n=1}^{N} \sum_{t=1}^{T}\frac{| y_{nt} - \hat y_{nt}|}{(|y_{nt}| + |\hat y_{nt}| + \epsilon) / 2}
\label{eq:smape}
\end{equation}

\begin{equation}
    \textsf{CORR} = \frac{\sum_{n=1}^{N} \sum_{t=1}^{T} (y_{nt} - \bar y)(\hat y_{nt} - \bar {\hat y})}{\sqrt{\sum_{n=1}^{N} \sum_{t=1}^{T} (y_{nt} - \bar y)^2} \sqrt{\sum_{n=1}^{N} \sum_{t=1}^{T} (\hat y_{nt} - \bar {\hat y})^2}}
\label{eq:corr}
\end{equation}

\subsection{Platform, Hardware and Limitations}
The AutoSeries challenge is hosted on CodaLab\footnote{\url{https://autodl.lri.fr/}}, an open sourced challenge platform. We provide  4-core 30GB memory CPU and no GPU is available. Participants may submit at most 5 times per day. A docker is provided\footnote{\url{https://hub.docker.com/r/vergilgxw/autotable}} for executing submissions and for offline development. Participants can also install external packages if necessary.

\subsection{Baseline}
\label{baseline}

To help participants get started, we provided a baseline method, which is simple but contains necessary modules in the processing pipeline. Many paticipants' submissions were derived from this baseline. In what follows, we decompose solutions (baseline and winning methods) into three modules: \textbf{\color{blue} feature engineering} (including time processing, numerical features, categorical features), \textbf{\color{orange}  model training} (including models used, hyperparameter tuning, ensembling) and \textbf{\color{teal}  update strategy} (including when and how to update models with the steaming test data). For the baseline, such modules include:
\begin{itemize}
\setlength\itemsep{0em}
    \item {\bf \color{blue}  Feature engineering.} 
Multiple calendar features are extracted from the time stamp: \emph{year}, \emph{month}, \emph{day}, \emph{weekday}, and \emph{hour}. Categorical variables (or strings) are hashed to unique integers.
No preprocessing is applied to numerical features.
    \item {\bf \color{orange} Model training.} A single LightGBM \cite{lightgbm} model is used. A LightGBM regressor is instantiated by predetermined hyperparameters and there is no hyperparameter tuning. 
    \item {\bf \color{teal}  Update strategy.} Since the test data comes in a streaming way, we need an update strategy to incorporate new test data and adjust our model. However, due to time limit on update procedure, we can't update too frequently. The update strategy used in baseline is simple. We split all test timestamps by 5 segments and for every segment, we retrain the lightGBM with old training data and new segment of test data.
\end{itemize}

\begin{table}[ht!]
\caption{ Answers to the 10 challenge question. All of them are tackled to certain extent. Orange checkmark means the solution is trivial, though answers the question.}
\label{tab:answer}
\centering
\resizebox{\textwidth}{!}{
\begin{tabular}{|l|c|l|}
\toprule
Question & Answered? & Comment  \\ 
\midrule
Q1 \bf Beyond autoregression & \greencheck & Features \{$x_t$\} are leveraged \\
\midrule
Q2 \bf Explainability & \orangecheck & LightGBM outputs feature importance \\
\midrule
Q3 \bf Multivariate/multiple time series & \greencheck & All training data is used to fit \\ \midrule
Q4 \bf Diversity of sampling rates & \greencheck & Multiple calendar features are extracted \\ \midrule
Q5 \bf Heterogeneous series length & \greencheck & Long format data facilitates the issue \\ \midrule
Q6 \bf Missing data & \orangecheck & Missing data is imputed by mean value\\ \midrule
Q7 \bf Data streaming & \greencheck & Models are retrained every few steps \\ \midrule
Q8 \bf Joint model and HP selection & \orangecheck & Randomized grid search is applied \\ \midrule
Q9 \bf Transfer/Meta Learning & \orangecheck & Metadata (size, IdNum) is considered \\ \midrule
Q10 \bf Hardware constraints & \greencheck & Model training time is recorded \\
\bottomrule
\end{tabular}}
\end{table}

\subsection{Results}

The AutoSeries challenge lasted one month and a half. We received over 700 submissions and more than 40 teams from both Academia (University of Washington, Nanjing University, etc.) and Industry (Oura, DeepBlue Technology, etc.), coming from various countries including China, United States, Singapore, Japan, Russia, Finland, etc. In the Feedback Phase\footnote{\url{https://autodl.lri.fr/competitions/149\#results}}, the top five participants are: \textbf{rekcahd, DeepBlueAI, DenisVorotyntsev, DeepWisdom, Kon} while in the Private Phase, the top five participants are: \textbf{DenisVorotyntsev, DeepBlueAI, DeepWisdom, rekcahd, bingo}. It can be seen that team \textbf{rekcahd} seems to overfit on the Feedback Phase (additional experiments are provided in Sec \ref{sec:overfitting}). All winners use LightGBM \cite{lightgbm} which is boosting ensemble of decision trees dominating most tabular challenges. Only 1st winner and 2nd winner implements hyperparameter tuning module which is really a key to successful generalisation in AutoSeries. We briefly summarize the solutions and provide a detailed account in Appendix.

\begin{itemize}
\setlength\itemsep{0em}
    \item {\bf \color{blue}  Feature engineering.}  Calendar features e.g \emph{year}, \emph{month}, \emph{day} were extracted from timestamp. Lag/shift and diff features added to original numerical features. Categorical features were encoded in various ways to integers. 
    \item {\bf \color{orange} Model training.}  Only linear regression models and LightGBM were used. Most participants used default or fixed hyperparameters. Only the first winner made use of HPO. The second winner optimized only the learning rate.  LightGBM provides built-in feature importance/selection. Model ensembling was obtained by weighting models based on their performance in the previous round. 
    \item {\bf \color{teal}  Update strategy.} All participants updated their models. The update period was either hard coded, computed as a fixed fraction of the time budget, or re-estimating on-the-fly, given remaining time.  
\end{itemize}

We verified (in Table \ref{tab:answer}) that the challenge successfully answered the ten questions we wanted addressed (see Section \ref{sec:intro}).

\section{Post Challenge Experiments}

This section presents systematic experiments, which consolidate some of our findings and extend them. We are particularly interested in verifying the generalisation ability of winning solutions on a larger number of tasks, and comparing them with open-sourced AutoSeries solutions. We also revisit some of our challenge design choices to provide guidelines for future challenges, including time budget limitations, and choice and number of datasets.

\subsection{Reproducibility}
First, we reproduce the solutions of the top four participants and the baseline methods, on the 10 datasets of the challenge (from both phases). In the AutoSeries challenge, we only used the \textsf{RMSE} (Eq \ref{eq:rmse}) for evaluation. For a more thorough comparison, we also include the \textsf{SMAPE} (Eq \ref{eq:smape}) here for calculating the relative error (which is particualrly useful when the ground truth target is small, e.g. in the case of sales). 
The results are shown in the Table \ref{tab:rmse}, Table \ref{tab:smape}, Table \ref{tab:time}, Figure \ref{fig:perf} and Figure \ref{fig:perf_all_improv}. We ran each method on each dataset for 10 times with different random seed. For simplicity, we use \emph{1st DV, 2nd DB, 3rd DW, 4th Rek} to denote solutions from top 4 winners. 

We can observe that clear improvements have been made by the top winners, compared to the baseline, and both \textsf{RMSE} and \textsf{SMAPE} are significantly reduced. From Figure \ref{fig:perf} we can further visualize that, while sometimes the winners' solutions are close in \textsf{RMSE}, their \textsf{SMAPE} are totally different, which implies the necessity of using multiple metrics for evaluation. 

\subsection{Overfitting and generalisation}
\label{sec:overfitting}

\begin{figure}
    \centering
    \begin{subfigure}[b]{0.475\textwidth}
        \centering
        \includegraphics[width=\textwidth]{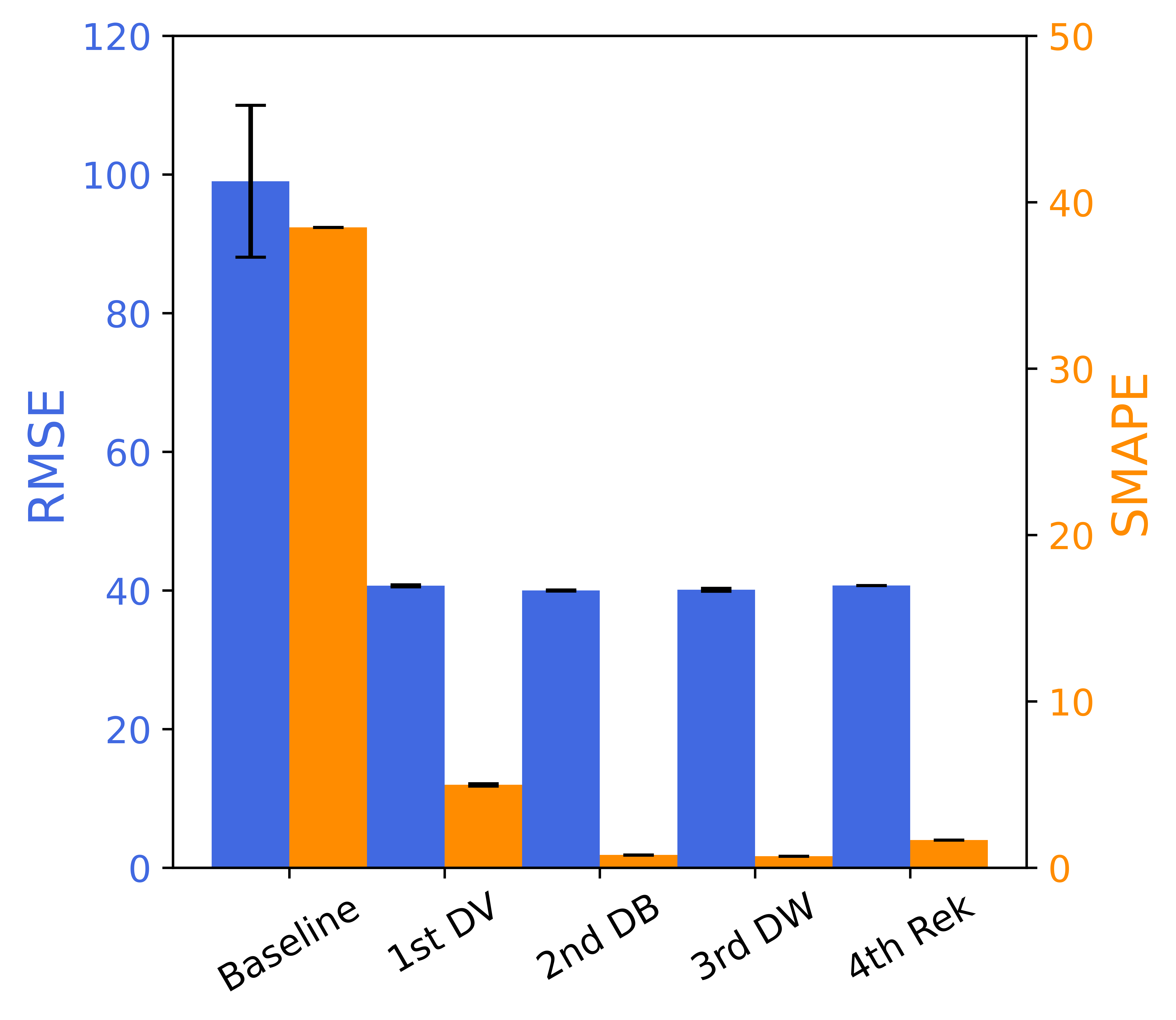}
        \caption[]{{\small Performance comparison on fph1}} 
        \label{fig:perf}
    \end{subfigure}
    % \hspace{0.1cm}
    % \vspace{-0.15cm}
    \begin{subfigure}[b]{0.475\textwidth} 
        \centering 
        \includegraphics[width=\textwidth]{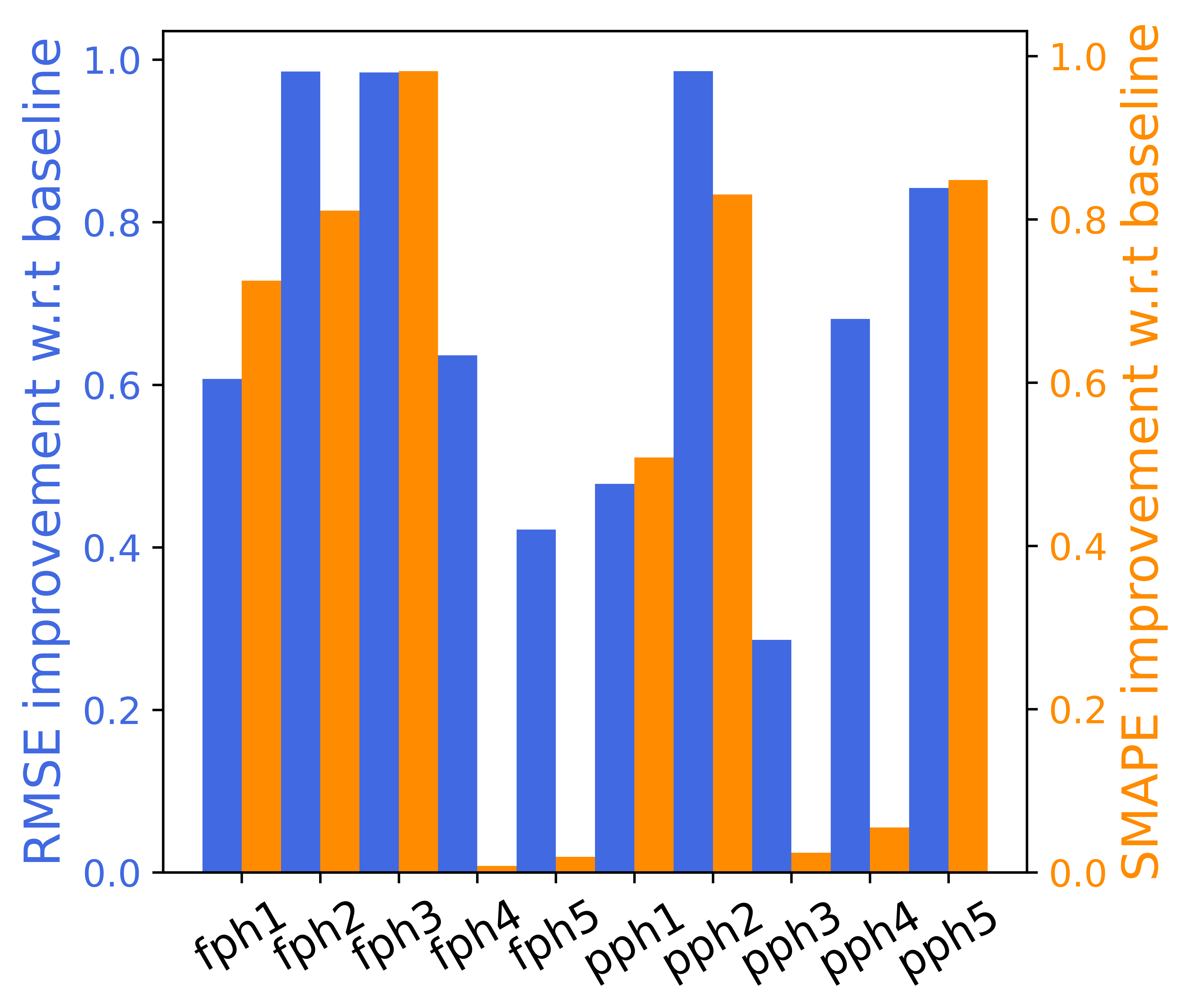}
        \caption[]{{\small Overall performance improvement }} 
        \label{fig:perf_all_improv}
    \end{subfigure}
    \hspace{-0.1\textwidth}
    \begin{subfigure}[b]{0.475\textwidth}   
        \centering 
        \includegraphics[width=\textwidth]{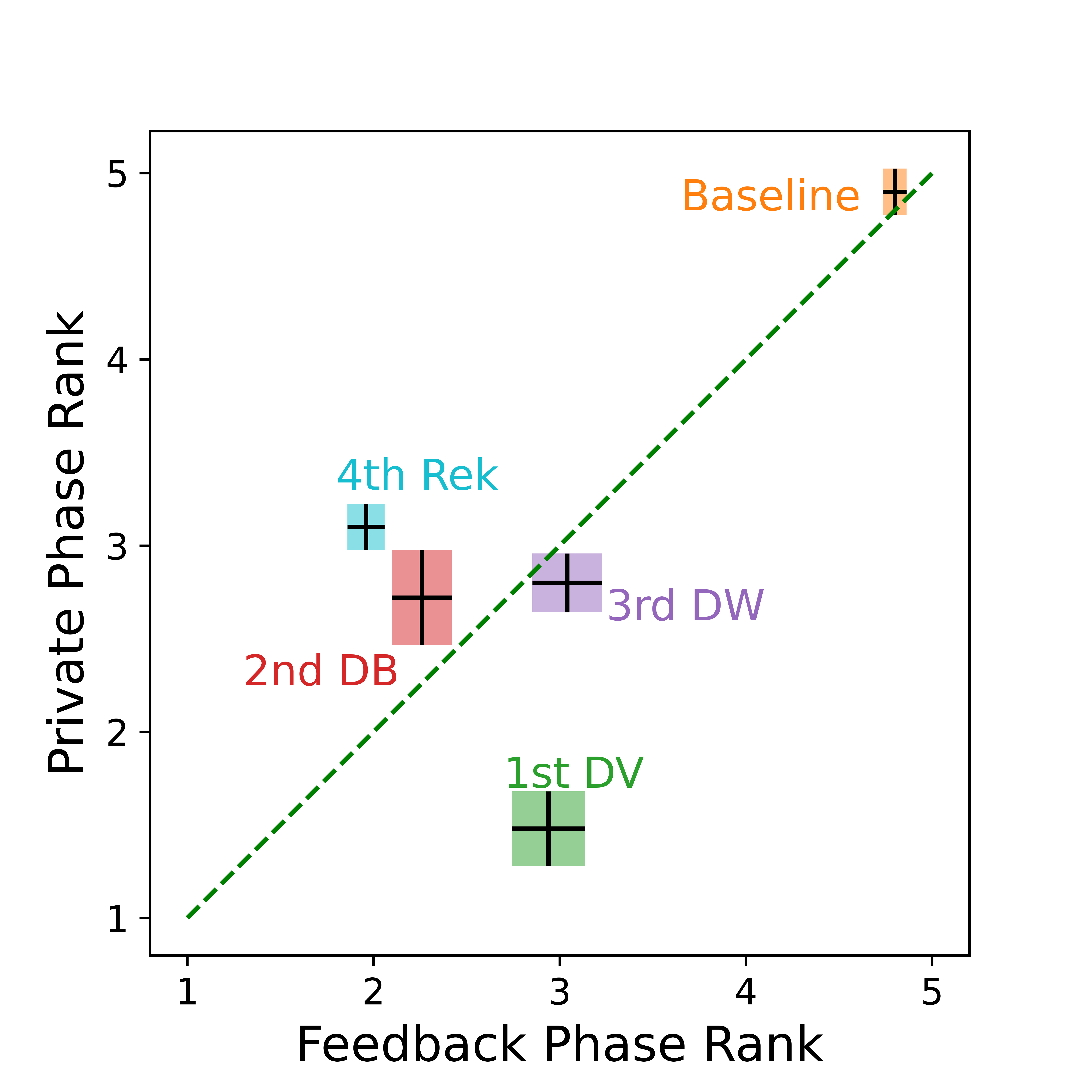}
        \caption[]{{\small Overfitting visualization}}
        \label{fig:overfitting}
    \end{subfigure}
    % \hspace{0.1\textwidth}
    \begin{subfigure}[b]{0.475\textwidth} 
        \centering 
        \includegraphics[width=\textwidth]{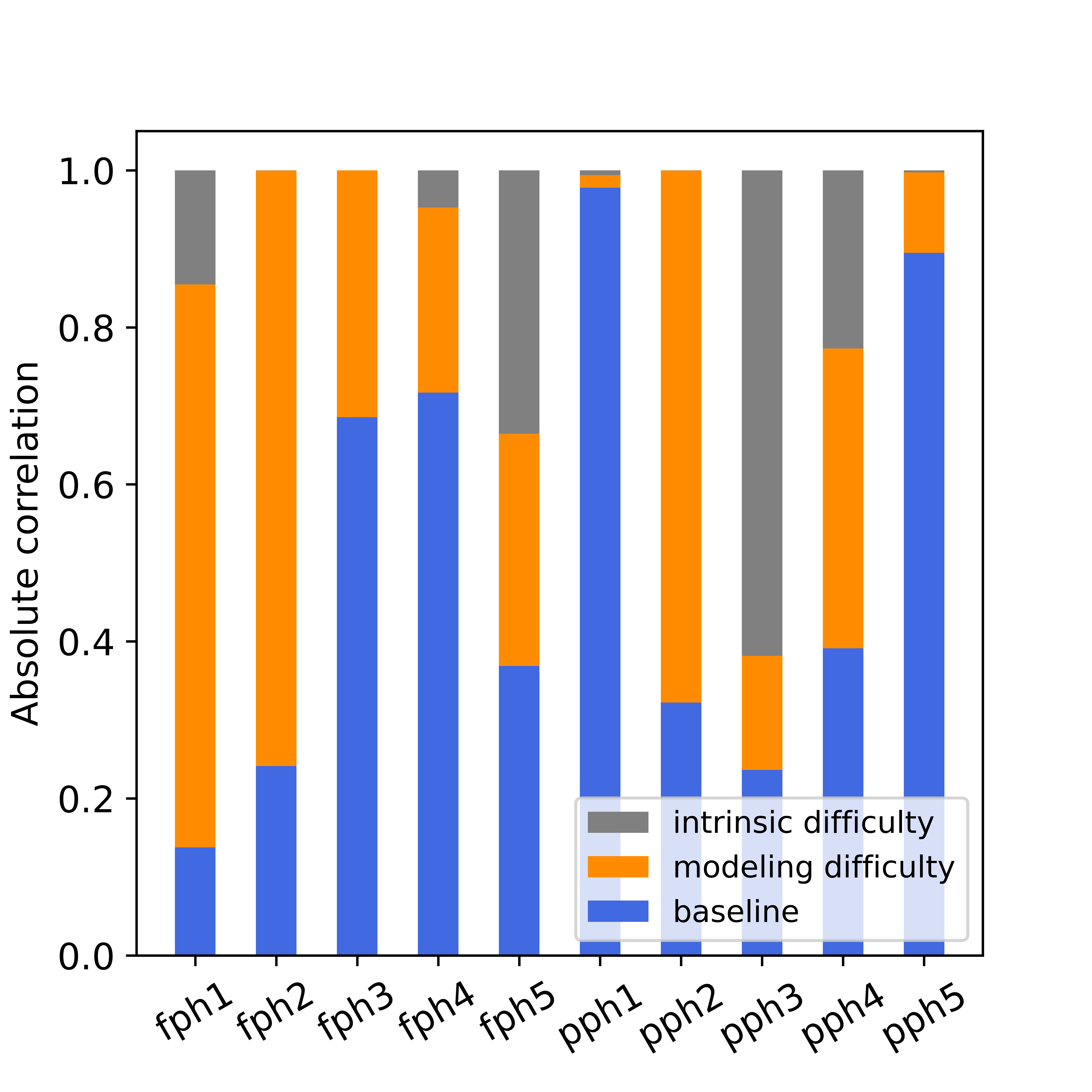}
        \caption[]{{\small Dataset difficulty}}    
        \label{fig:difficulty}
    \end{subfigure}
    \caption{\footnotesize {\bf Post challenge experiments.} \textbf{(a)} {\bf Performance comparison on dataset fph1.}
   We compare \textsf{RMSE} and \textsf{SMAPE} of all solutions on dataset fph1. Performances in \textsf{RMSE} are significantly better than the baseline for all winning teams, but all winners perform similarly. In contrast the \textsf{SMAPE} metric differentiates the winners, which focuses more on relative error. \textbf{(b)} {\bf Performance improvement on all datasets.} Both \textsf{RMSE} and \textsf{SMAPE} errors from best methods are compared to the baseline performance. The improvement ratio is calculated by (baseline score - best score) / baseline score. \textbf{(c)} {\bf Did the participants overfit the feed-back phase tasks?} Rankings are based on 10 runs: for each run, we rank separately in order to validate the stability of methods. Regions show the mean and std of rankings over multiple runs. Methods in the upper triangle are believed to overfit, e.g. 4th winner's solution. Methods in the lower triangle are believed to generalize well, e.g. 1st winner's solution. \textbf{(d)} {\bf How well did we choose the 10 datasets?} As explained in Sec \ref{sec:difficulty}, we use absolute correlation as a bounded metric for calculating a notion of intrinsic difficulty (gray bar) and modeling difficulty (orange bar) of the 10 datasets. Datasets with high modeling difficulty and low intrinsic difficulty are better choices in a benchmark.}
    \label{fig:postexp}
\end{figure}

Based on our reproduced results, we analyse potential overfitting which is visualized in Figure \ref{fig:overfitting}. Among each run (based on a different random seed), we rank solutions on feedback phase datasets and private phase datasets separately. Rankings are based on \textsf{RMSE} as in AutoSeries challenge. After 10 runs, we plot the mean and std of the ranking as a region. This shows that 4th Rek overfits to feedback datasets since it performs very well in feedback phase but poorly in private phase. But it is also interesting to visualize that 1st DV has a good generalisation: although it is not the best in feedback phase, it achieves great results in private phase. Including hyperparameter search may have provided the winner with a key advantage.

\begin{table}
     \caption{ {\bf Post-challenge runs}. We repeated 10 times all runs on all datasets for the top ranking submissions of the private phase. Each run is based on a different random seed. Error bar are indicated (one standard deviation) unless no variance was observed (algorithm with no stochastic component, such as 4th Rek).\\}
    \begin{subtable}{\textwidth}
        \centering
        \resizebox{0.85\textwidth}{!}{
        \begin{tabular}{|c|c|c|c|c|c|c|} 
        \toprule
        Dataset & Phase & Baseline & 1st DV & 2nd DB & 3rd DW & 4th Rek \\ 
        \midrule
        \rowcolor{green!40} fph1 & Feedback &  100$\pm$10 &  40.7$\pm$0.2 & {\bf 40.0$\pm$0.1}	& 40.1$\pm$0.2 & 40.7 \\ 
        \rowcolor{green!40} fph2 & Feedback & 18000$\pm$2000 &	237$\pm$2 &	244$\pm$1  &	243.9$\pm$0.3 &	{\bf 230.7}\\ 
        \rowcolor{green!40} fph3 & Feedback & 3000 & 600$\pm$20 & 53.04$\pm$0.01 & {\bf 52.4} & 108.4 \\ 
        \rowcolor{green!40} fph4 & Feedback & 6.9$\pm$0.3	 &3.66$\pm$0.02 &	2.760$\pm$0.007	 & NA  &	{\bf 2.632} \\ 
        \rowcolor{green!40} fph5 & Feedback & 8.6$\pm$0.7& 5.76$\pm$0.01 & 5.760$\pm$0.005 & 5.780$\pm$0.003 & {\bf 5.589} \\ 
        \midrule
        \rowcolor{yellow!60} pph1 & Private & 400$\pm$6 &	{\bf 200$\pm$2} &	223.5$\pm$0.7&	200$\pm$10	&420.8 \\ 
        \rowcolor{yellow!60} pph2 & Private & 17000$\pm$4000	& {\bf 240$\pm$2} &	253.7$\pm$0.5&	260$\pm$2&	246.7 \\ 
        \rowcolor{yellow!60} pph3 & Private & 9$\pm$3&	{\bf 6.20$\pm$0.02}&	6.330$\pm$0.007&	6.40$\pm$0.03&	6.56 \\
        \rowcolor{yellow!60} pph4 & Private & 12$\pm$2&	4.0$\pm$0.2&	3.80$\pm$0.04	&3.700$\pm$	0.004&	{\bf 3.32} \\ 
        \rowcolor{yellow!60} pph5 & Private & 300$\pm$30&	{\bf 50$\pm$1} &	100	&60$\pm$20 &	167.8 \\ 
        \bottomrule
        \end{tabular}
        }
       \caption{\footnotesize \textsf{RMSE} comparison.}
       \label{tab:rmse}
    \end{subtable}
    \hfill
    \begin{subtable}{\textwidth}
        \centering
        \resizebox{0.85\textwidth}{!}{
        \begin{tabular}{|c|c|c|c|c|c|c|} 
        \toprule
        Dataset & Phase & Baseline & 1st DV & 2nd DB & 3rd DW & 4th Rek \\ 
        \midrule
        \rowcolor{green!40} fph1 & Feedback & 140$\pm$20 &  100.0$\pm$0.5 & 104.00$\pm$0.07	& 104.00$\pm$0.04 & {\bf 40.77} \\ 
        \rowcolor{green!40} fph2 & Feedback & 140$\pm$10&	{\bf 30$\pm$1} &	40.0$\pm$0.3  &	38.8$\pm$0.1&	33.9\\ 
        \rowcolor{green!40} fph3 & Feedback & 38.49 & 5.0$\pm$0.1 & 0.770$\pm$0.001& {\bf 0.700$\pm$0.001} & 1.674\\ 
        \rowcolor{green!40} fph4 & Feedback & 190$\pm$1	 &191.0$\pm$0.1&	191.0$\pm$0.1	 & NA  &	{\bf 186.2} \\ 
        \rowcolor{green!40} fph5 & Feedback & {\bf 170$\pm$1} & 173.5$\pm$0.1 & 174.1$\pm$	0.1 & 172.9$\pm$0.1 & 170.6 \\ 
        \midrule
        \rowcolor{yellow!60} pph1 & Private & 12.7$\pm$0.6&	{\bf 6.1$\pm$0.2} &	6.49$\pm$0.03&	6.4$\pm$0.8 & 12.14\\ 
        \rowcolor{yellow!60} pph2 & Private & 140$\pm$10 & {\bf 24.0$\pm$0.5} & 35.0$\pm$0.3 &	31.0$\pm$0.1 &	32.75 \\ 
        \rowcolor{yellow!60} pph3 & Private & 180$\pm$3 &	180.0$\pm$0.7&	181.0$\pm$0.1&	180$\pm$0.1& {\bf 174.7}\\ 
        \rowcolor{yellow!60} pph4 & Private & 170$\pm$3&	170$\pm$1&	167.8$\pm$0.1 	& 170.0$\pm$0.2 &	{\bf 164}\\ 
        \rowcolor{yellow!60} pph5 & Private & 40$\pm$2&	{\bf 6.0$\pm$0.5} &	9	& 8$\pm$3 &	30.61\\ 
        \bottomrule
        \end{tabular}
        }
        \caption{\footnotesize \textsf{SMAPE} comparison.}
        \label{tab:smape}
     \end{subtable}
\end{table}

\begin{table}
\caption{ Supported features comparison between various open-source packages and the AutoSeries winning solution (also open-sourced).}
\label{tab:features}
\centering
\resizebox{\textwidth}{!}{
\begin{tabular}{|c|c|c|c|c|} 
 \toprule
 Solutions & FeatureEngineering & ModelTraining & StreamingUpdate & TimeManagement  \\ 
 \midrule
 Featuretools\cite{kanter_dfs} & Tabular & \redcross & \redcross & \redcross \\
%  \midrule
 tsfresh\footnote{https://github.com/blue-yonder/tsfresh} & Temporal & \redcross & \redcross & \redcross \\
%  \midrule
 Prophet\cite{fbprophet} & \redcross & \greencheck & \redcross & \redcross \\
%  \midrule
 GluonTS\cite{gluonts} & Temporal & \greencheck & \redcross & \redcross \\
%  \midrule
 AutoKeras\cite{autokeras} & \redcross & \greencheck & \redcross & \greencheck \\
%  \midrule
 AutoGluon\cite{autogluon} & Tabular & \greencheck & \redcross & \greencheck \\
%  \midrule
 Google AutoTable\footnote{\url{https://cloud.google.com/automl-tables}} & Tabular & \greencheck & \greencheck & \greencheck \\
 \midrule
 \bf AutoSeries & \bf Temporal & \greencheck & \greencheck & \greencheck \\
 \bottomrule
\end{tabular}
}
\end{table}

\subsection{Comparison to open source AutoML solutions}

In this section, we turn our attention to comparing AutoSeries with similar open-source solutions. However, to the best of our knowledge, there is no publicly available AutoML framework dedicated to time series data. Current features (categorized by three modules of solutions as in Sec ) of open source packages, which can be used to tackle the problems of the challenge with some engineering effort, are summarized in Table \ref{tab:features}.

Packages like Featuretools, tsfresh focus on (tabular, temporal) feature engineering; they do not provide trainable models and should be used in conjuction with another package.Prophet and GluonTS are known for rapid prototyping with time series, but they are not AutoML packages (in the sense that they do not come with automated model selection and hyper-parameter selection). AutoKeras is an package focusing more on image and text, with KerasTuner\footnote{\url{https://keras-team.github.io/keras-tuner/}} for neural architecture search. Google AutoTable meets most of our requirements, but is not open sourced, and is not dedicated to time series. Moreover, Google AutoTable costs around 19 dollars per hour in order to train on 92 computing instances at the same time, which is far more than our challenge settings.

\begin{figure}
    \centering
    \begin{subfigure}[b]{0.45\textwidth}
        \centering
        \includegraphics[width=0.95\textwidth]{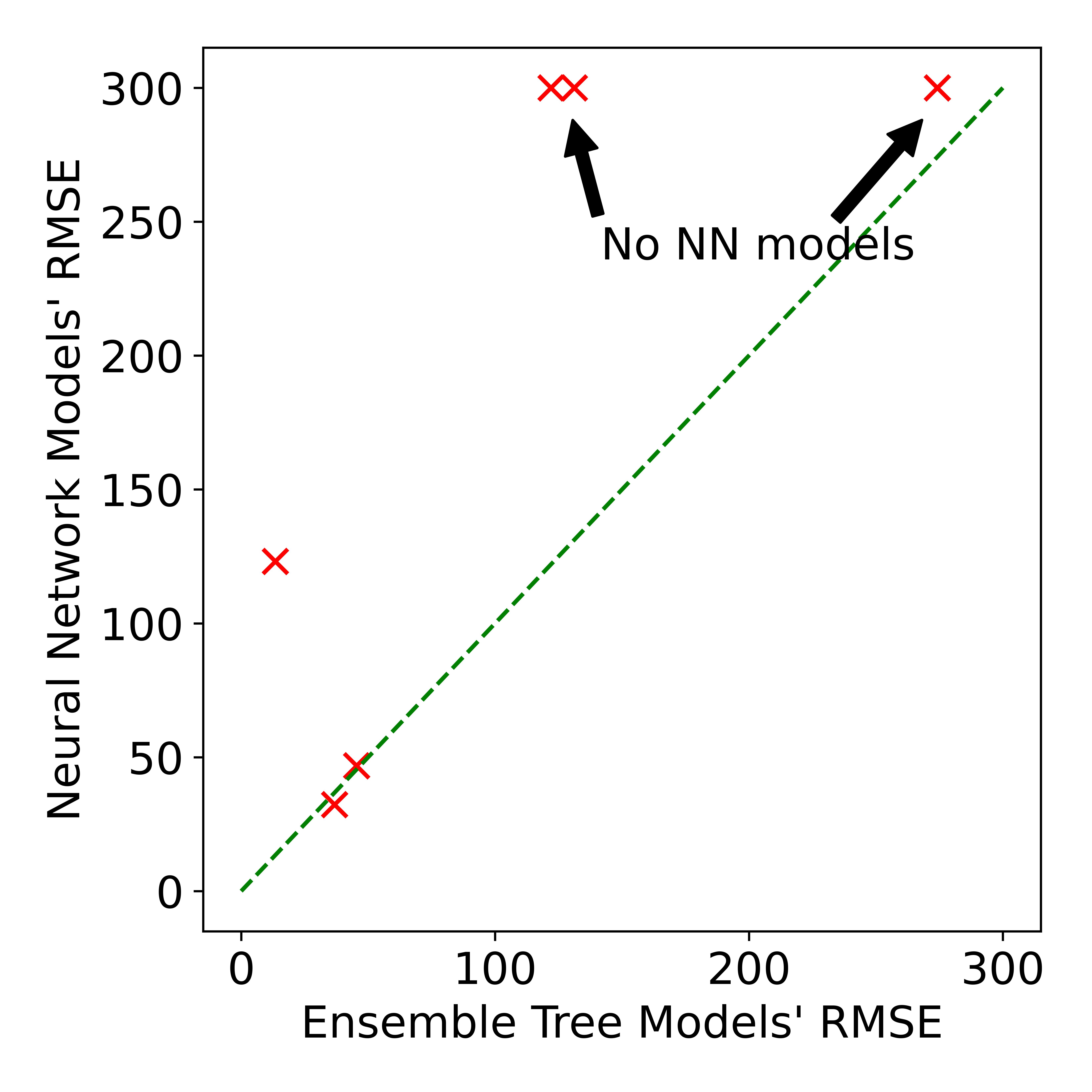}
        \caption[]{{\small AutoGluon algorithm groups}} 
        \label{fig:algogroup}
    \end{subfigure}
    % \vspace{-0.15cm}
    \begin{subfigure}[b]{0.45\textwidth} 
        \centering 
        \includegraphics[width=\textwidth]{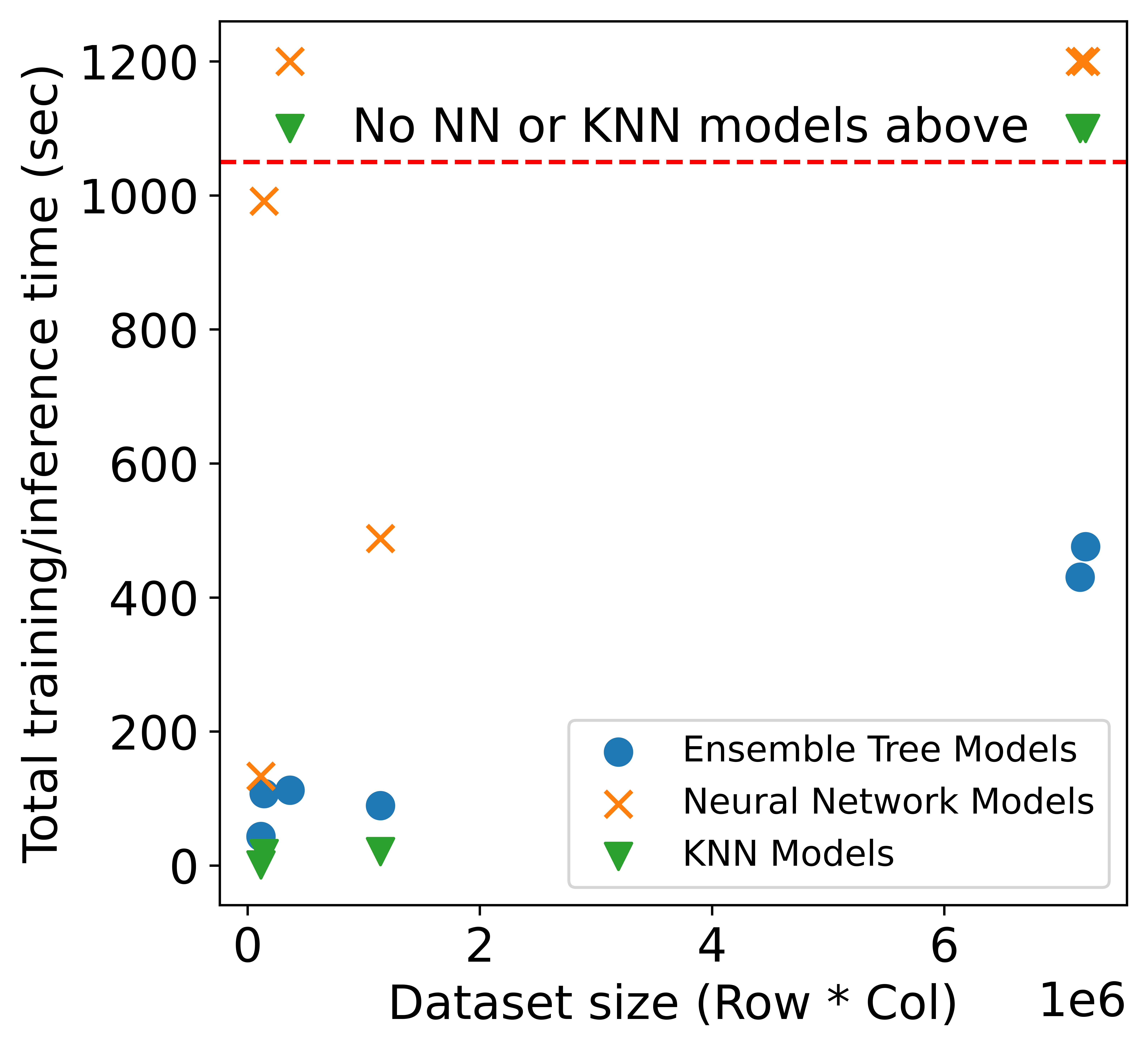}
        \caption[]{{\small Time-Size by groups }} 
        \label{fig:timesize}
    \end{subfigure}
    % \hspace{0.1\textwidth}
    \caption{{\bf AutoGluon experiments.} \textbf{(a)} {\bf Average performance for two algorithm groups.} Here we compare the average \textsf{RMSE} of Neural Network models and Ensemble Tree models. Among the six feasible datasets for AutoGluon with time series feature engineering,  3 of them don't choose Neural Networks as ensemble candidates. Ensemble trees have always significantly better performances. On 2 datasets, Neural Networks are quite competitive. \textbf{(b)} {\bf Average time costs of candidate models.} When dataset is large, only ensemble tree models are chosen. When dataset is medium, KNN is fastest, followed by tree models. Neural Networks take significantly more time.}
    \label{fig:ag}
\end{figure}

At last, we selected AutoGluon for comparison, as being closest to our use case. AutoGluon provides end-to-end automated pipelines to handle tabular data without any human intervention (e.g. hyperparameter tuning, data preprocessing). AutoGluon includes many more candidate models and fancier ensemble methods than the wining solutions, but its feature engineering is not dedicated to multivariate time series. For example, it doesn't distinguish time series Id combinations to summarize statistics of one particular time series. We ran AutoGluon on all 10 datasets with default parameters except for using \textsf{RMSE} as evaluation metric and best\_quality as presets parameter. The results are summarized in Table \ref{tab:ag} column AutoGluon. Not surprisingly, vanilla AutoGluon can only beat the baseline, and it is significantly worse than the winning solutions. We further compile AutoGluon with 1st winner's time series feature engineering and update the models the same way as in baseline. The results are in Table \ref{tab:ag} column FE+AutoGluon. AutoGluon can now indeed achieve comparable results with best winner and sometimes even better, which strongly implies the importance of time series feature engineering. Note that we didn't limit strictly AutoGluon's running time as in our challenge. In general, AutoGluon takes 10 times more time than the winning solution and it still can't output a valid performance on the four datasets in a reasonable time. For the six AutoGluon's feasible datasets, we further visualize in Figure \ref{fig:ag} by algorithm groups. AutoGluon contains mainly three algorithm groups: Neural Network (MXNet, FastAI), Ensemble Trees (LightGBM, Catboost, XGBoost) and K-Nearest Neighbors. We first plot on the left the average \textsf{RMSE} for Neural Networks models and ensemble tree models each (we omit KNN methods since they are usually the worst). Note that among the 6 datasets, 3 datasets don't use Neural Network for final ensemble (so their \textsf{RMSE} are set to be a large number for visualization). On 2 datasets (bottom left corner), however, Neural Networks can be competitive. This encourages us to explore in the future the effectiveness of deep models on time series which evolve quickly these days. On the right, we average the training/inference time per algorithm group and find that KNN can be used for very fast prediction if needed. Neural Networks take significantly more time. Points above the dotted line mean that no NN models or KNN models are chosen for this dataset (either due to performance or time cost). Only the tree-based methods provide solutions across the range of dataset sizes.

\begin{table}[]
\caption{ {\bf Comparison with AutoGluon.} NA means a missing value: AutoGluon did not terminate within a reasonable time.}
\label{tab:ag}
\centering
\resizebox{\textwidth}{!}{
\begin{tabular}{|c|c|cc|cc|cc|cc|}
\toprule
Dataset & Phase & \multicolumn{2}{c|}{Baseline} & \multicolumn{2}{c|}{1st DV} & \multicolumn{2}{c|}{AutoGluon} & \multicolumn{2}{c|}{FE + AutoGluon} \\
\midrule
& & \textsf{RMSE} & \textsf{SMAPE} & \textsf{RMSE} & \textsf{SMAPE} & \textsf{RMSE} & \textsf{SMAPE} & \textsf{RMSE} & \textsf{SMAPE} \\
\midrule
\rowcolor{green!40} fph1 & Feedback &  99.04 & 142.59 & 40.69 & 102.19 & 90.19 & {\bf 26.45} & {\bf 40.57} & 105.31 \\ 
\rowcolor{green!40} fph2 & Feedback & 17563 & 142.64 & {\bf 236.6} & 26.63 & 14978 & 59.94 & 263.74 & {\bf 25.51} \\ 
\rowcolor{green!40} fph3 & Feedback & 3337 & 38.49 & {\bf 623.32} & {\bf 4.99} & 6365 & 116.14 & 3159 & 31.08 \\ 
\rowcolor{green!40} fph4 & Feedback & 6.91 & {\bf 187.58} & {\bf 3.66} & 190.94 & NA & NA & NA & NA\\ 
\rowcolor{green!40} fph5 & Feedback & 8.63 & 174.45 & {\bf 5.76} & {\bf 173.54} & NA & NA & NA & NA\\ 
\midrule
\rowcolor{yellow!60} pph1 & Private & 422.37 & 12.65 & 218.83 & 6.11 & 2770.70 & 9.46 & {\bf 212.68} & {\bf 5.85} \\
\rowcolor{yellow!60} pph2 & Private & 16851 & 139.31 & {\bf 242.41} & 23.46 & 15028 & 57.04 & 269.85 & {\bf 22.98} \\ 
\rowcolor{yellow!60} pph3 & Private & 8.78 & 178.45 & {\bf 6.21} & {\bf 177.08} & NA & NA & NA & NA\\ 
\rowcolor{yellow!60} pph4 & Private & 11.54 & 174.94 & {\bf 3.74} & {\bf 168.4} & NA & NA & NA & NA\\ 
\rowcolor{yellow!60} pph5 & Private & 309.33 & 39.2 & {\bf 50.37} & {\bf 5.91} & 949.4 & 20.52 &  65.22 & 6.65 \\ 
\bottomrule
\end{tabular}
}
\end{table}

\subsection{Impact of time budget}
\label{sec:budget}

In the AutoSeries challenge, time management is an important aspect. Different time budgets are allowed for different datasets (as shown in Table \ref{tab:stats}). Ideally, AutoSeries solutions should take into account the allowed time budget and adapt all modules in the pipeline (i.e. different feature engineering, model training and updating strategy based on different allowed time budgets). We double the time budget and compare the performance in Appendix. In general, no obviously stable improvement can be observed. We also try to half the time budget and most solutions can't even produce valid predictions meaning that no single model training is finished. This could be because that we set the defaults budgets too tight but it also shows from another perspective that participants' solutions overfit to the challenge design (default time budget).

\subsection{Dataset Difficulty}
\label{sec:difficulty}

After a challenge finishes, another important issue for the organizers is to validate the choice of datasets. This is particularly interesting for AutoML challenges since the point is to generalize to a wide variety of tasks. Inspired by difficulty measurements in \cite{liu_autocv}, we want to define intrinsic difficulty and modeling difficulty. By intrinsic difficulty we mean the irreducible error.
As a surrogate to the intrinsic difficulty, we use the error of the best model. 
By modeling difficulty, we mean the range or spread of performances of candidate models. To separate well competition participants, we want to choose datasets of low intrinsic difficulty and high modeling difficulty. In \cite{liu_autocv}, a notion of intrinsic difficulty and modeling difficulty is introduced for classification problems. Here we adapt such ideas and choose another bounded metric, the correlation (\textsf{CORR}) (Eq \ref{eq:corr}). 
In fact, correlation has been used in many time series papers as a metric \cite{lai_lstnet,wang_defsi}. We calculate the absolute correlation between the prediction sequence and ground truth test sequence. We define {\bf Intrinsic difficulty} as $1$ minus the best solution’s absolute correlation score; and  {\bf Modeling difficulty} as the difference between the best solution’s absolute correlation score and the provided baseline score.

 These difficulty measures are visualized in Figure \ref{fig:difficulty}. It is obvious that both intrinsic difficulty and modeling difficulty differ from datasets to datasets. A posteriori, we can observe that some datasets like pph1 and pph5 are too easy, while pph3 is too difficult. In general, feedback datasets are of higher quality than private datasets, which is unfortunate. However, it is also possible that participants overfit the feedback datasets and thus, by using the best performing methods to estimate the intrinsic difficulty, we obtain a biased estimation. 

\section{Conclusion and future work}

In this challenge, we introduce an AutoML setting with streaming test data, aiming at pushing forward research on Automated Time Series, and also having an impact on industry. Since there were no open sourced AutoML solutions dedicated to time series prior to our challenge, we believe the open sourced AutoSeries solutions fill this gap and provide a useful tool to researchers and practitioners.  AutoSeries solutions don't need a GPU which facilitates their adoption.

The solutions of the winners are based on lightGBM. They addressed all challenge questions, demonstrating the feasibility of automating time series regression on datasets of the type considered. Significant improvements were made compared to the provided baseline. Our generalisation and overfitting experiments show that hyperparameter search is key to generalize. 
Still, some of the questions were addressed in a rather trivial way and deserve further attention. Explainabilty boils down to the feature importance delivered by lightGBM. In future challenge designs, we might want to quantitatively evaluate this aspect. Missing data were trivially imputed with the mean value. Hyper-parameters were not thoroughly optimized by most participants, and simple random search was used (if at all). Our experiments with the AutoGluon package demonstrate that much can be done in this direction to further improve results. Additionally, no sophisticated method of transfer learning or meta-learning was used. Knowledge transfer was limited to the choice of features and hyper-parameters performed on the feedback phase datasets. New challenge designs could include testing meta-learning capabilities on the platform, by letting the participant's code meta-train on the platform, e.g. not resetting the model instances when presented with each new dataset.

Other self criticisms of our design include that some datasets in the private phase may have been too easy or too difficult. Additionally, the \textsf{RMSE} alone could not separate well solutions, while a combination of metrics might be more revealing. Lastly, GPUs were not provided. On one hand this forced the participants to deliver practical rapid solutions; on the other hand, this precluded them from exploring neural time series models, which are rapidly progressing in this field. 

Finally, winning solutions overfitted to the provided time budgets (no improvement with more time and fail with less time). An incentive to encourage participants to deliver ``any-time-learning'' solutions as opposed to ``fixed-time-learning'' solutions is to use the area under the learning curve as metric, as we did in other challenges. We will consider this for future designs.

\section*{Acknowledgments}
Funding and support have been received by several research grants, including ANR Chair of Artificial Intelligence HUMANIA ANR-19-CHIA-00222-01, Big Data Chair of Excellence FDS Paris-Saclay, Paris Région Ile-de-France, 4Paradigm, ChaLearn, Microsoft, Google. We would like also to thank the following people for their efforts in organizing AutoSeries challenge, insightful discussions, etc., including Xiawei Guo, Shouxiang Liu, Zhenwu Liu.

\bibliography{ref}

\begin{thebibliography}{16}
\providecommand{\natexlab}[1]{#1}
\providecommand{\url}[1]{\texttt{#1}}
\expandafter\ifx\csname urlstyle\endcsname\relax
  \providecommand{\doi}[1]{doi: #1}\else
  \providecommand{\doi}{doi: \begingroup \urlstyle{rm}\Url}\fi

\bibitem[Alexandrov et~al.(2020)Alexandrov, Benidis, Bohlke{-}Schneider,
  Flunkert, Gasthaus, Januschowski, Maddix, Rangapuram, Salinas, Schulz,
  Stella, T{\"{u}}rkmen, and Wang]{gluonts}
Alexander Alexandrov, Konstantinos Benidis, Michael Bohlke{-}Schneider,
  Valentin Flunkert, Jan Gasthaus, Tim Januschowski, Danielle~C. Maddix,
  Syama~Sundar Rangapuram, David Salinas, Jasper Schulz, Lorenzo Stella,
  Ali~Caner T{\"{u}}rkmen, and Yuyang Wang.
\newblock Gluonts: Probabilistic and neural time series modeling in python.
\newblock \emph{J. Mach. Learn. Res.}, 2020.

\bibitem[Erickson et~al.(2020)Erickson, Mueller, Shirkov, Zhang, Larroy, Li,
  and Smola]{autogluon}
Nick Erickson, Jonas Mueller, Alexander Shirkov, Hang Zhang, Pedro Larroy,
  Mu~Li, and Alexander Smola.
\newblock Autogluon-tabular: Robust and accurate automl for structured data.
\newblock 2020.

\bibitem[Hutter et~al.(2018)Hutter, Kotthoff, and Vanschoren]{automl_book}
Frank Hutter, Lars Kotthoff, and Joaquin Vanschoren, editors.
\newblock \emph{Automated Machine Learning: Methods, Systems, Challenges}.
\newblock Springer, 2018.
\newblock http://automl.org/book.

\bibitem[Hyndman and Athanasopoulos(2021)]{fpp3_book}
Rob~J. Hyndman and George Athanasopoulos, editors.
\newblock \emph{Forecasting: principles and practice}.
\newblock OTexts, 2021.
\newblock OTexts.com/fpp3. Accessed on 2021/03/25.

\bibitem[Jin et~al.(2019)Jin, Song, and Hu]{autokeras}
Haifeng Jin, Qingquan Song, and Xia Hu.
\newblock Auto-keras: An efficient neural architecture search system.
\newblock In \emph{KDD}, 2019.

\bibitem[Kanter and Veeramachaneni(2015)]{kanter_dfs}
James~Max Kanter and Kalyan Veeramachaneni.
\newblock Deep feature synthesis: Towards automating data science endeavors.
\newblock In \emph{{IEEE} International Conference on Data Science and Advanced
  Analytics, DSAA}, 2015.

\bibitem[Ke et~al.(2017)Ke, Meng, Finley, Wang, Chen, Ma, Ye, and
  Liu]{lightgbm}
Guolin Ke, Qi~Meng, Thomas Finley, Taifeng Wang, Wei Chen, Weidong Ma, Qiwei
  Ye, and Tie-Yan Liu.
\newblock Lightgbm: A highly efficient gradient boosting decision tree.
\newblock In \emph{Advances in Neural Information Processing Systems}, 2017.

\bibitem[Lai et~al.(2018)Lai, Chang, Yang, and Liu]{lai_lstnet}
Guokun Lai, Wei{-}Cheng Chang, Yiming Yang, and Hanxiao Liu.
\newblock Modeling long- and short-term temporal patterns with deep neural
  networks.
\newblock In \emph{SIGIR}, 2018.

\bibitem[Lim and Zohren(2020)]{lim_tsf}
Bryan Lim and Stefan Zohren.
\newblock Time series forecasting with deep learning: A survey.
\newblock 2020.

\bibitem[Liu et~al.(2020)Liu, Xu, Escalera, Guyon, J{\'{u}}nior, Madadi, Pavao,
  Treguer, and Tu]{liu_autocv}
Zhengying Liu, Zhen Xu, Sergio Escalera, Isabelle Guyon, J{\'{u}}lio C.
  S.~Jacques J{\'{u}}nior, Meysam Madadi, Adrien Pavao, S{\'{e}}bastien
  Treguer, and Wei{-}Wei Tu.
\newblock Towards automated computer vision: analysis of the autocv challenges
  2019.
\newblock \emph{Pattern Recognit. Lett.}, 2020.

\bibitem[Prokhorenkova et~al.(2018)Prokhorenkova, Gusev, Vorobev, Dorogush, and
  Gulin]{catboost}
Liudmila~Ostroumova Prokhorenkova, Gleb Gusev, Aleksandr Vorobev, Anna~Veronika
  Dorogush, and Andrey Gulin.
\newblock Catboost: unbiased boosting with categorical features.
\newblock In \emph{Advances in Neural Information Processing Systems}, 2018.

\bibitem[Tan et~al.(2020)Tan, Bergmeir, Petitjean, and Webb]{tan_tsr}
Chang~Wei Tan, Christoph Bergmeir, Francois Petitjean, and Geoffrey~I. Webb.
\newblock Time series extrinsic regression.
\newblock 2020.

\bibitem[Taylor and Letham(2017)]{fbprophet}
Sean~J. Taylor and Benjamin Letham.
\newblock Forecasting at scale.
\newblock \emph{PeerJ Prepr.}, 2017.

\bibitem[Wang et~al.(2019)Wang, Chen, and Marathe]{wang_defsi}
Lijing Wang, Jiangzhuo Chen, and Madhav Marathe.
\newblock {DEFSI:} deep learning based epidemic forecasting with synthetic
  information.
\newblock In \emph{AAAI}, 2019.

\bibitem[Wang et~al.(2017)Wang, Yan, and Oates]{wang_tsc}
Zhiguang Wang, Weizhong Yan, and Tim Oates.
\newblock Time series classification from scratch with deep neural networks:
  {A} strong baseline.
\newblock In \emph{International Joint Conference on Neural Networks}, 2017.

\bibitem[Yao et~al.(2018)Yao, Wang, Chen, Dai, Li, Tu, Yang, and
  Yu]{yao_automl}
Quanming Yao, Mengshuo Wang, Yuqiang Chen, Wenyuan Dai, Yu-Feng Li, Wei-Wei Tu,
  Qiang Yang, and Yang Yu.
\newblock Taking human out of learning applications: A survey on automated
  machine learning.
\newblock 2018.

\end{thebibliography}

\newpage

\begin{appendix}
\section{Detailed descriptions of winning methods}\label{appendix A}

\subsection{First place: DenisVorotyntsev}

The 1st winning is from team DenisVorotyntsev. Their code is open sourced on GitHub\footnote{\url{https://github.com/DenisVorotyntsev/AutoSeries}}.\\

\noindent
{\bf \color{blue} Feature engineering.} First, a small LightGBM model is fit on training data and the top 3 most important numerical features are extracted. Then pairwise arithmetic operations are conducted on these top candidates to generate extra numerical features. Afterwards, a large number of lag features are generated. To deal with multivariate time series which contain one or more ID columns for indexing time series, a \emph{batch\_id} column is created by concatenating all ID columns (as strings). This \emph{batch\_id} column will be used by \emph{groupby} in further processing steps. 

For generating lag/shift features on target column, data is grouped by \emph{batch\_id} and for each group (i.e. one particular time series dataframe), lagged/shifted target and differences with respect to lagged/shifted target are recorded as additional features. Lags are by default a list of small number e.g. 1,2,3,5,7. Same lag feature generation is performed on numerical features as well, with fewer lags e.g. 1,2,3.

Categorical columns are concatenated with corresponding time series \emph{batch\_id}. They are encoded by CatBoost \cite{catboost}. The Target is linearly transformed to have minimum equal to 1. A possible target difference operation is optional and is added to the hyperparameter search space. 
Calendar features are extracted as in the baseline method.\\

\noindent
{\bf \color{orange} Model training.} LightGBM is the only used model. For the first time training, three steps are performed: base model fit, feature selection, hyperparameter search. For a base model, a  lightGBM model is fit on all features including generated feature columns. Then a feature selection step is performed to remove least important features. They search in order 5 different splits \emph{0.2, 0.5, 0.75, 0.05, 0.1} of most important features and fit on the selected ones. The best performing ratio is recorded. Lastly, a relatively large hyperparameter search space is defined and in total 2880 configurations of hyperparameters are randomly searched under training time limit. These hyperparameters include \emph{num\_leaves}, \emph{min\_child\_samples}, \emph{subsample\_freq}, \emph{colsample\_bytree}, \emph{subsample}, \emph{lambda\_l2}. \\

\noindent
{\bf \color{teal} Update strategy.} The core of update strategy is to determine the update frequency. They first calculate the affordable training time from the training time budget and the first training time cost, with a coefficient buffer. Then the update frequency is simply calculated with the number of test steps divided by affordable training rounds. After the update frequency is determined, the update function is called per certain steps, which calls the training module with the training data and newly streamed data. 

\subsection{Second place: DeepBlueAI}

The second winning solution is from the team DeepBlueAI. Their code is open sourced on GitHub\footnote{\url{https://github.com/DeepBlueAI/AutoSeries}}.\\

\noindent
{\bf \color{blue} Feature engineering.} A mean imputer is used for missing data. Calendar features including \emph{year}, \emph{month}, \emph{day}, \emph{weekday}, \emph{hour}, \emph{minute} are extracted. In the case of multivariate time series, ID columns are merged into a unique identifier. Categorical features are encoded by pandas Categorical. For numerical features, many more features are generated after grouping by unique ID: mean, max, min, lag, division, etc. No feature selection is applied. \\

\noindent
{\bf \color{orange} Model training.} Two models are used: linear regression and LightGBM. For linear regression, they distinguish data with few features (less than 20) and data with rich features. In the former case, a simple fit is used. When there are lots of features, they select features based on a F-score of regression. For LightGBM, most hyperparameters are predefined except for learning rate. They choose learning rate by fitting different LightGBM models with different hyperparameters according to dataset size. What's particular in the training and predict part is that they maintain two coefficients for ensembling linear model and LightGBM model. These coefficients are searched by comparing to ground truth.\\

\noindent
{\bf \color{teal} Update strategy.} The update strategy is the same as in the baseline method.

\subsection{Third place: DeepWisdom}

The third winning solution is from the team DeepWisdom. Their code is open sourced on GitHub\footnote{\url{https://github.com/DeepWisdom/AutoSeries2019}}.\\

\noindent
{\bf \color{blue} Feature engineering.} Four types of features are generated sequentially: KeysCross, TimeData, Delta and LagFeat. KeysCross applies in the case of multiple ID\_Key for indexing time series, for example in the retail setting, we may have shop\_id and item\_id together to index the time series for certain item in certain shop. KeysCross converts string ID columns to integers first and then for each ID columns $c$, existing ID encoding is multiplied by $c.max$ and is added by $c.values$. TimeData is the calendar feature. It extracts \emph{year, month, day, hour, weekday} from the timestamp column. Both Delta and LagFeat deal with the target column. They are basically diff and lag/shift features. Delta calculates first order difference, second order difference and numerical exchange ratio of the target. LagFeat calculates \emph{mean, std, max, min} for a lag window of size \emph{3,7,14,30}. For linear regression models used in later steps (as mentioned in next paragraph), one hot encoding of categorical variables and a mean imputer for numerical variables are used. \\

\noindent
{\bf \color{orange} Model training.} Three models are used for training: lasso regression, ridge regression and LightGBM. For all these models, hyperparameters are predefined manually and there is no hyperparameter search. During the prediction step, all trained models will produce outputs and a weighted linear combination is used for ensembling these models. The weights are inversely proportional to a model's previous performance of \textsf{RMSE}.\\

\noindent
{\bf \color{teal} Update strategy.} The update strategy is the same as that of the 1st Place DenisVorotyntsev.

\subsection{Fourth place: Rekcahd}

The fourth place participant is from team rekcahd. Their code isn't open sourced since this team is not among top three, and thus it was not a requirement for them to publish their solution. This solution relies a lot on the provided baseline and it achieved first place in the Feedback Phase.\\

\noindent
{\bf \color{blue} Feature engineering.} Missing values are filled by feature means. Calendar features are extracted as in baseline Sec \ref{baseline}. The type adaptation module deals with categorical features particularly. In each group of a categorical feature (i.e. a sub dataframe where this particular category of same value. Note that all time series exist in the group), this categorical value (as long as there are not too few instances in the group), is mapped to a linear combination of target mean of this group and target mean of the whole training data.

For multivariate time series datasets, they add extra  features. Concretely, two shifted target features are added (shifted once and twice); if target is always positive, square root of the target is added; difference with respect to shifted once values are calculated for numerical features; another feature for each categorical feature indicating whether it is same as previous time's value is added. However, these shift, difference and category change indicator features are not calculated on univariate time series datasets, which explains to some extent the higher error of this solution on two univariate time series (fph3, pph1).\\

\noindent
{\bf \color{orange} Model training.} A linear regression model is fit on training data and is served as starting score for a LightGBM model. All hyperparameters of LightGBM are predefined except for \emph{num\_boost\_round}, which is determined by the best iteration after fitting another LightGBM model.\\

\noindent
{\bf \color{teal} Update strategy.} The update strategy is the same as in baseline.

\newpage
\section{Dataset difficulty based on other metrics}\label{appendix B}

\begin{figure}[H]
    \centering
    \begin{subfigure}[b]{0.475\textwidth}
        \centering
        \includegraphics[width=\textwidth]{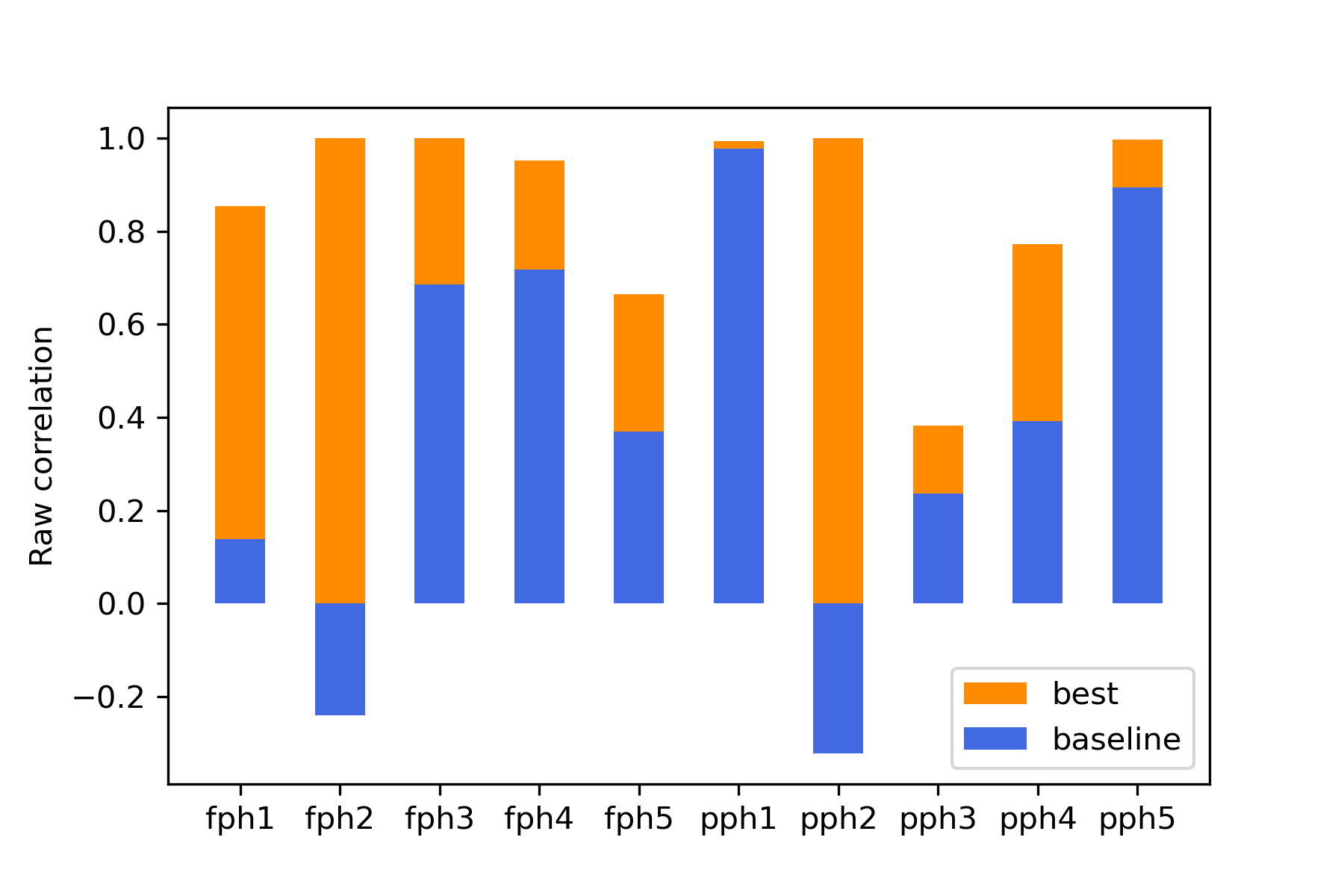}
        \caption{\footnotesize Raw corr difficulty measure of 10 datasets.} 
        \label{fig:difficulty_corr}
    \end{subfigure}
    % \hspace{0.1cm}
    \vspace{-0.15cm}
    \begin{subfigure}[b]{0.475\textwidth} 
        \centering 
        \includegraphics[width=1\linewidth]{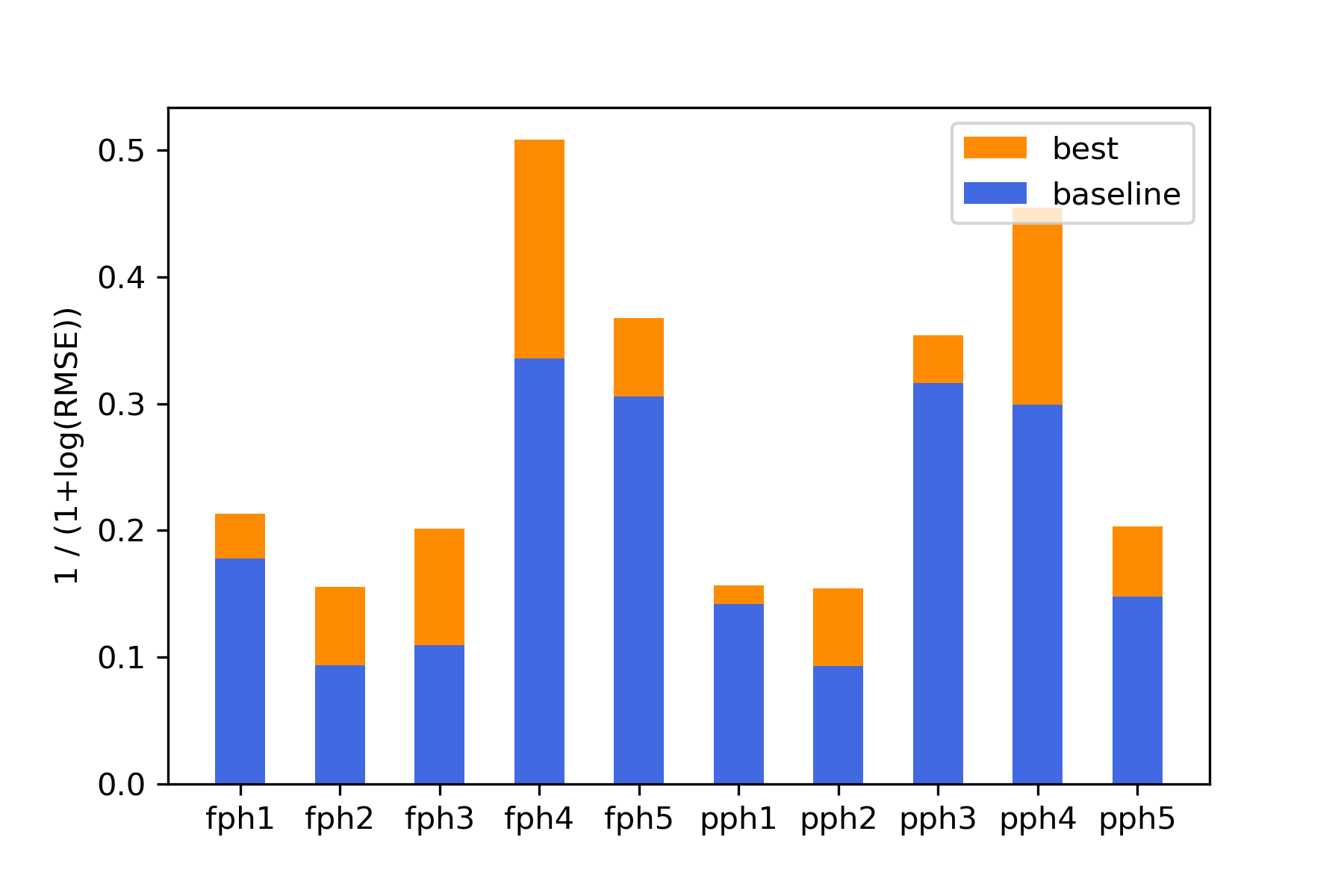}
        \caption{\footnotesize RMSE difficulty measure of 10 datasets.}
        \label{fig:difficulty_rmse} 
    \end{subfigure}
    \hspace{-0.1\textwidth}
    \begin{subfigure}[b]{0.475\textwidth}   
        \centering 
        \includegraphics[width=1\linewidth]{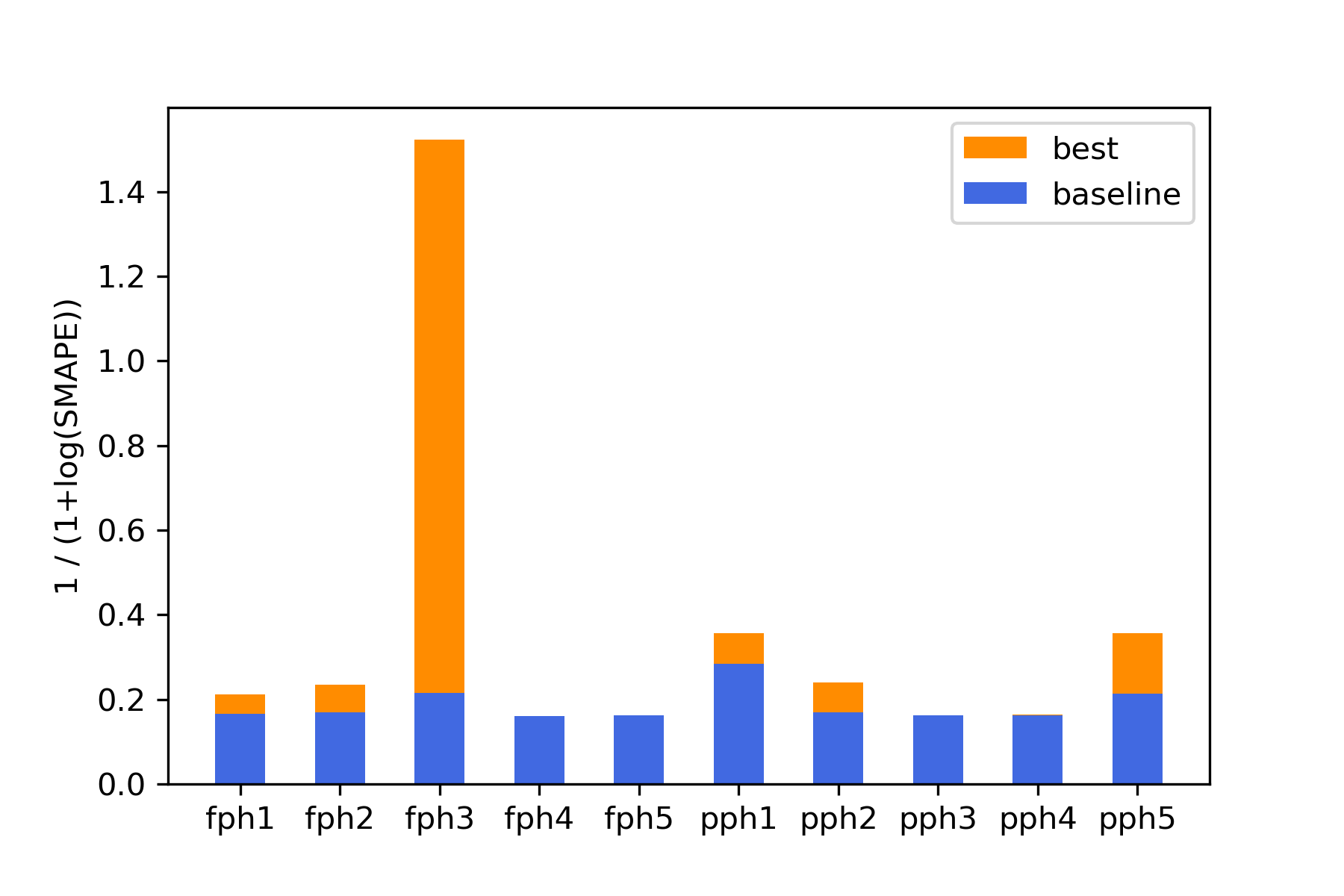}
        \caption{\footnotesize SMAPE difficulty measure of 10 datasets.}
        \label{fig:difficulty_smape} 
    \end{subfigure}
    % \hspace{0.1\textwidth}
    \begin{subfigure}[b]{0.475\textwidth} 
        \centering 
        \includegraphics[width=1\linewidth]{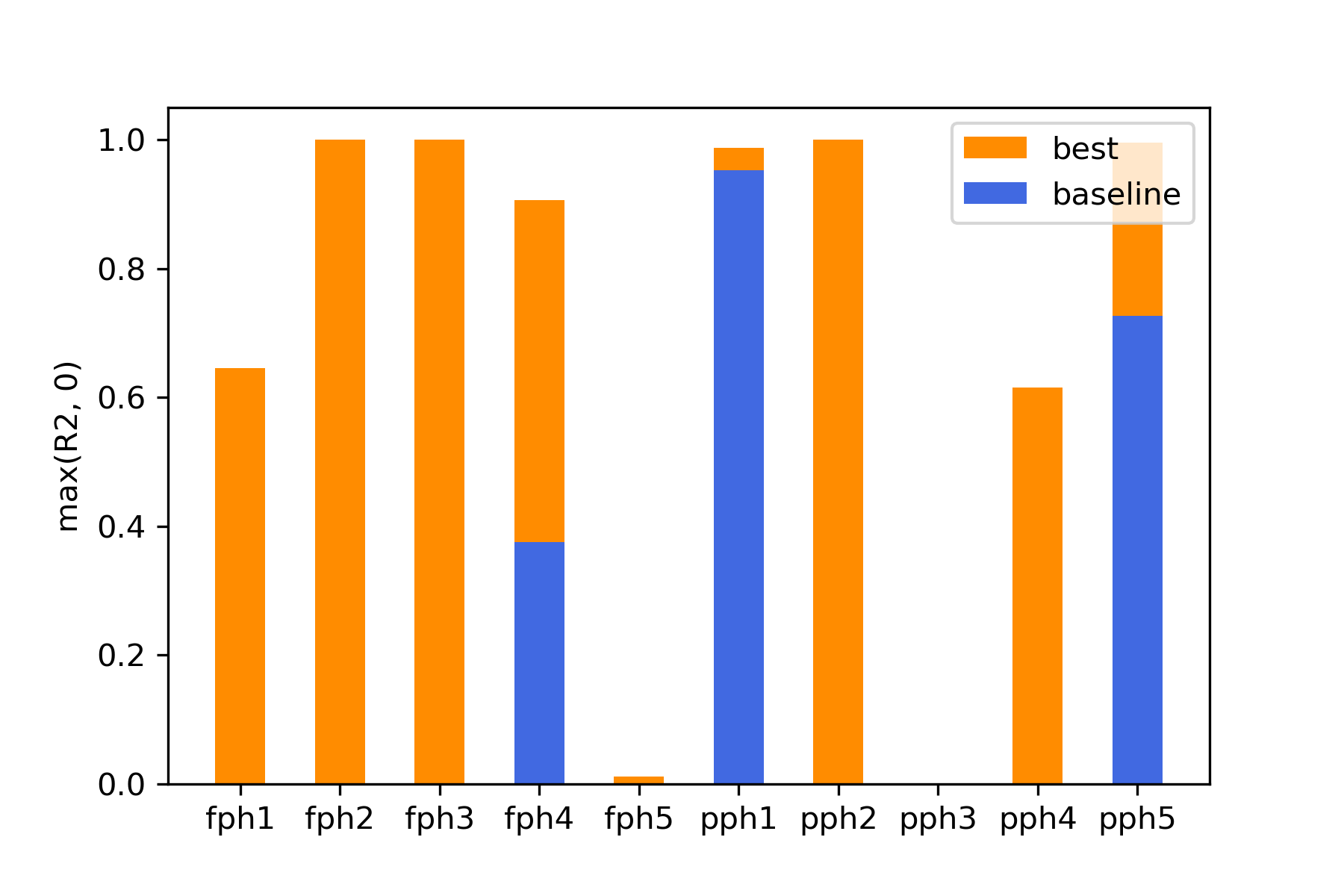}
        \caption{\footnotesize R2 difficulty measure of 10 datasets.}
        \label{fig:difficulty_r2} 
    \end{subfigure}
    \caption{\bf Dataset difficulty based on other metrics.}
\end{figure}

\newpage
\section{Running time comparison}\label{appendix C}

\begin{table}[H]
\caption{{\bf Performance improvement with double time budget.} A positive number (in bold) means performances have been improved (in percentage) given more time.}
\label{tab:doubletime}
\centering
\resizebox{\textwidth}{!}{
\begin{tabular}{@{}|cc|cc|cc|cc|cc|cc|@{}}
\toprule
Dataset & Phase & \multicolumn{2}{c|}{Baseline} & \multicolumn{2}{c|}{1st DV} & \multicolumn{2}{c|}{2nd DB}  & \multicolumn{2}{c|}{3rd DW} & \multicolumn{2}{c|}{4th Rek}\\
\midrule
& & \textsf{RMSE} & \textsf{SMAPE} & \textsf{RMSE} & \textsf{SMAPE} & \textsf{RMSE} & \textsf{SMAPE} & \textsf{RMSE} & \textsf{SMAPE} & \textsf{RMSE} & \textsf{SMAPE}\\
\midrule
\rowcolor{green!40} fph1 & Feedback & {\bf 23.50} & {\bf 18.24} & -0.02 & -1.47 & {\bf 0.05} & 0.00 & -0.18 & 0.00 & 0.00 & 0.00 \\
\rowcolor{green!40} fph2 & Feedback & {\bf 26.70} & {\bf 3.84} & -1.57 & {\bf 1.76} & -0.90 & -1.61 & {\bf 78.53} & {\bf 98.17} & 0.00 & 0.00 \\
\rowcolor{green!40} fph3 & Feedback & 0.00 & 0.00 & {\bf 0.65} & {\bf 0.50} & -0.04 & 0.00 & 0.00 & 0.00 & 0.00 & 0.00 \\
\rowcolor{green!40} fph4 & Feedback & {\bf 2.59} & -1.39 & {\bf 0.35} & {\bf 0.05} & {\bf 0.04} & 0.00 & NA & NA & 0.00 & 0.00 \\
\rowcolor{green!40} fph5 & Feedback & -15.63 & {\bf 1.43} & -0.38 & {\bf 0.06} & -0.05 & {\bf 0.06} & 0.00 & 0.00 & 0.00 & 0.00 \\
\midrule
\rowcolor{yellow!60} pph1 & Private & -1.11 & -3.65 & {\bf 1.17} & {\bf 7.97} & 0.00 & {\bf 0.03} & -14.54 & -43.05 & 0.00 & 0.00 \\
\rowcolor{yellow!60} pph2 & Private &{\bf 47.32} & {\bf 18.08} & {\bf 1.35} & -4.05 & {\bf 0.35} & {\bf 0.57} & {\bf 0.61} & {\bf 0.92} & 0.00 & 0.00 \\
\rowcolor{yellow!60} pph3 & Private &-13.18 & -1.72 & {\bf 1.23} & {\bf 0.11} & -0.27 & -0.17 & NA & NA & {\bf 59.90} & -6.58 \\
\rowcolor{yellow!60} pph4 & Private &-119.24 & {\bf 5.67} & -1.23 & {\bf 0.24} & -0.21 & -0.06 & NA & NA & 0.00 & 0.00 \\
\rowcolor{yellow!60} pph5 & Private &-6.98 & -13.37 & {\bf 2.97} & -16.32 & 0.00 & 0.00 & {\bf 37.44} & {\bf 35.26} & 0.00 & 0.00 \\
\bottomrule
\end{tabular}
}
\end{table}

\begin{table}[H]
\caption{ {\bf Duration comparison.} Mean and Std calculated on 10 runs. (Unit: sec). Each run is the same setting except for a different random seed. The method consuming most allowed time is bolded.}
\label{tab:time}
\centering
\resizebox{\textwidth}{!}{
\begin{tabular}{|c|c|c|c|c|c|c|c|} 
 \toprule
 Dataset & Phase &Budget & Baseline & 1st DV & 2nd DB & 3rd DW & 4th Rek \\ 
 \midrule
 \rowcolor{green!40} fph1 & Feedback & 1300 & 70$\pm$10 &  {\bf 700$\pm$100} & 600$\pm$100	& 600$\pm$50 & 260$\pm$60 \\ 
 \rowcolor{green!40} fph2 & Feedback & 2000 & 150$\pm$20&	1200$\pm$100&	550$\pm$80 & {\bf 1300$\pm$70}&	200$\pm$30\\ 
 \rowcolor{green!40} fph3 & Feedback & 500 & 16$\pm$2&150$\pm$10& 70$\pm$10& {\bf 160$\pm$20} & 14$\pm$3\\ 
 \rowcolor{green!40} fph4 & Feedback & 3500 & 600$\pm$60 &{\bf 1200$\pm$40}&	1000$\pm$80	 & NA  &	800$\pm$70 \\ 
 \rowcolor{green!40} fph5 & Feedback & 2000 & 200$\pm$20& 900$\pm$30 & 900$\pm$80 & {\bf 1100$\pm$50} & 470$\pm$40\\ 
\midrule
 \rowcolor{yellow!60} pph1 & Private & 1600 & 100$\pm$20 &	900$\pm$150 &	{\bf 900$\pm$200} &	890$\pm$70 & 150$\pm$30 \\ 
 \rowcolor{yellow!60} pph2 & Private & 2000 & 150$\pm$15 &1100$\pm$100 & 600$\pm$80&	{\bf 1400$\pm$50} & 170$\pm$30\\ 
 \rowcolor{yellow!60} pph3 & Private & 3500 & 300$\pm$30 & 1300$\pm$40&	900$\pm$80 &	{\bf 2000$\pm$60} & 550$\pm$50 \\ 
 \rowcolor{yellow!60} pph4 & Private & 2000 & 450$\pm$50 &	860$\pm$60&	870$\pm$100	& {\bf 1000$\pm$80} &	500$\pm$50 \\ 
 \rowcolor{yellow!60} pph5 & Private & 350 & 15$\pm$2 &	{\bf 200$\pm$8}&	140$\pm$20	& 170$\pm$6 &	26$\pm$4\\ 
 \bottomrule
\end{tabular}
}
\end{table}

\end{appendix}

\end{document}